\documentclass[letterpaper]{article}
\usepackage[preprint]{aaai2027}
\usepackage[hyphens]{url}
\usepackage{graphicx}
\usepackage{natbib}
\usepackage{caption}
\usepackage{amsmath}
\usepackage{amsfonts}
\usepackage{amssymb}
\usepackage{booktabs}
\usepackage{xcolor}
\usepackage{algorithm}
\usepackage{algpseudocode}
\usepackage{verbatim}

\definecolor{loopflagblue}{RGB}{31,96,180}

\newcommand{\projectpagelink}{%
  \leavevmode
  \pdfstartlink attr{/Border [0 0 0]} user{%
    /Subtype /Link /A << /S /URI
    /URI (https://yuanyiyan.com/projects/looped-flow-matching) >>}%
  \textcolor{loopflagblue}{\textbf{Project Page}}%
  \pdfendlink
}

\newcommand{\firstpageteaser}{%
  \vspace{-0.4em}
  \begin{center}
  \includegraphics[width=0.92\textwidth]{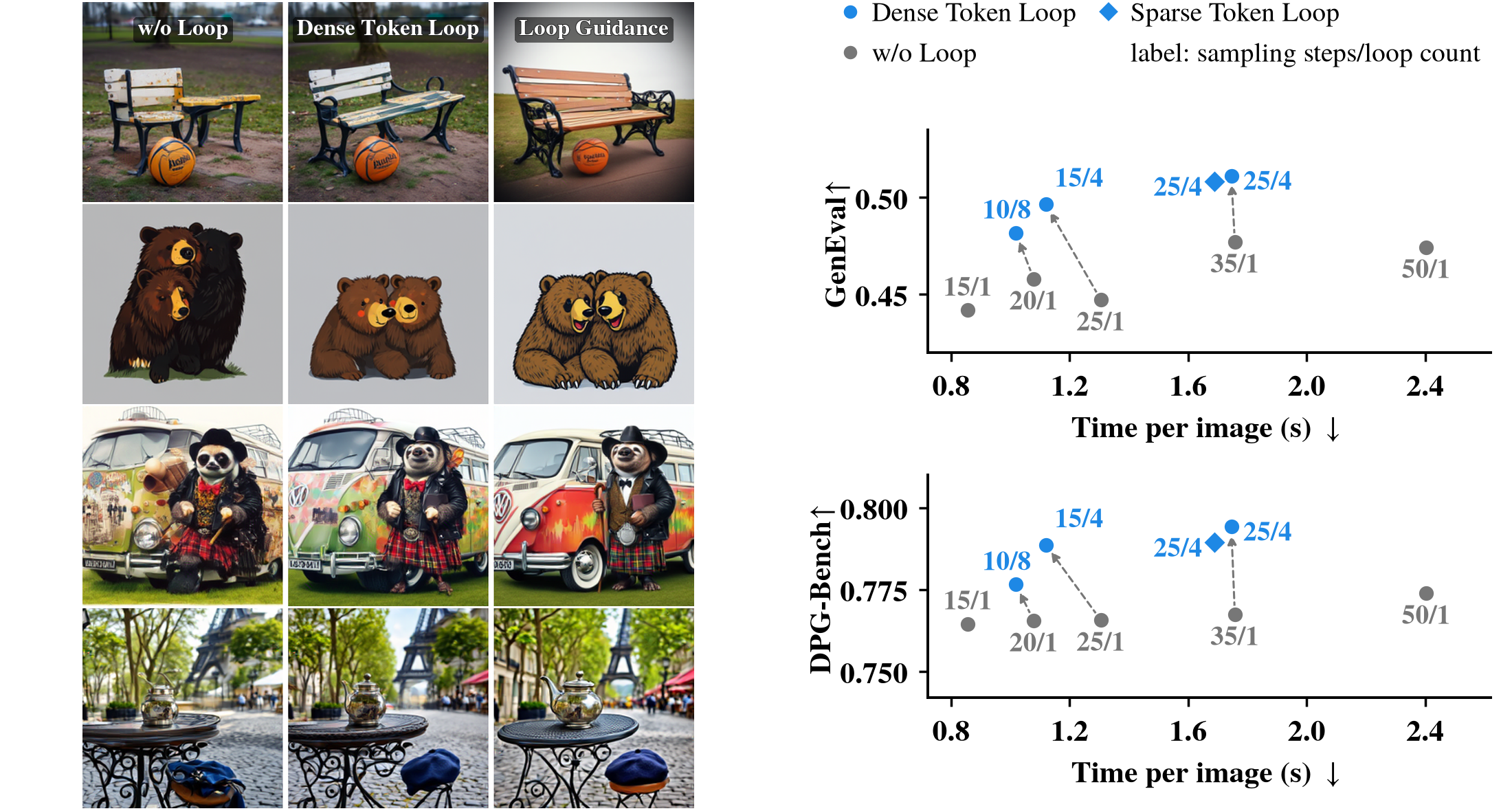}
  \captionof{figure}{Looping inside the denoiser produces cleaner images and
  better quality--time trade-offs. Left: matched Scale-RAE prompts compare the
  no-loop baseline, Dense Token Loop, and Loop Guidance. Right: Dense and
  Sparse Token Loop improve GenEval and DPG-Bench over no-loop baselines with
  more outer sampling steps, at comparable or lower generation time; point
  labels denote sampling steps over loop count.}
  \label{fig:qualitative_main}
  \end{center}
  \vspace{0.5em}
}

\makeatletter
\def\maketitle{%
  \par%
  \begingroup
    \def\thefootnote{\fnsymbol{footnote}}%
    \twocolumn[\@maketitle\firstpageteaser]%
    \long\def\@footnotetext##1{\insert\aaai@thanksins{%
        \protect\footnotesize\interlinepenalty\interfootnotelinepenalty
        \splittopskip\footnotesep\splitmaxdepth\dp\strutbox
        \floatingpenalty\@MM\hsize\columnwidth\@parboxrestore
        \protected@edef\@currentlabel{%
           \csname p@footnote\endcsname\@thefnmark}%
        \color@begingroup
          \@makefntext{%
            \rule\z@\footnotesep\ignorespaces##1\@finalstrut\strutbox}%
        \color@endgroup}}%
    \@thanks%
  \endgroup%
  \if T\copyright@on\insert\aaai@copyrightins{\noindent\footnotesize\copyright@text}\fi%
  \setcounter{footnote}{0}%
  \let\maketitle\relax%
  \let\@maketitle\relax%
  \gdef\@thanks{}%
  \gdef\@author{}%
  \gdef\@title{}%
  \let\thanks\relax%
}%
\makeatother

\title{Training-Free Hidden-State Refinement for Flow-Matching Image Generators}
\author{
Yuanyi Yan\textsuperscript{\rm 1},
Xinzhe Rao\textsuperscript{\rm 1},
Canyu Shen\textsuperscript{\rm 2},
Yang Chen\textsuperscript{\rm 1}\\
Yunlu Chen\textsuperscript{\rm 3},
Meng Tang\textsuperscript{\rm 4},
Teng Long\textsuperscript{\rm 5},
Vincent Tao Hu\textsuperscript{\rm 1}
}

\affiliations{
\textsuperscript{\rm 1}Huazhong University of Science and Technology\\
\textsuperscript{\rm 2}Tongji University\\
\textsuperscript{\rm 3}King Abdullah University of Science and Technology\\
\textsuperscript{\rm 4}University of California, Merced\\
\textsuperscript{\rm 5}University of Amsterdam\\[0.35em]
\projectpagelink
}

\begin{document}

\maketitle

\begin{abstract}
We aim to improve frozen flow-matching image generators by adding inference
computation inside the denoiser, without changing model weights or the outer
sampler. Existing generators usually spend extra test-time computation by
increasing the number of sampling steps, which repeatedly evaluates the entire
denoiser and couples quality gains to sampler cost. A key challenge is how to
use extra computation inside a frozen transformer denoiser: the method must
decide which tokens, layers, and sampling times receive repeated updates while
preserving the original generation pipeline. We introduce a training-free
looping framework that repeatedly applies selected transformer layers inside
each denoising call. Dense and Sparse Token Loop vary the token scope;
Sampling-Progress Gating and the loop layer range specify when and where looping
is active; loop count and strength control the repeated updates; and Loop
Guidance combines ordinary and looped vector-field predictions. Across two
Scale-RAE model scales, loop variants improve primary and auxiliary quality
metrics with competitive quality--efficiency trade-offs. Loop Guidance further
improves both primary metrics across all three tested models; on Scale-RAE
DiT2.4B, it raises GenEval from 0.4471 to 0.5691 and DPG-Bench from 0.7656 to
0.8053. Code will be released.
\end{abstract}

% ----- Inlined from 0_intro.tex -----
\section{Introduction}

Flow matching has become a central formulation for modern generative modeling
because it learns a time-dependent vector field that transports a simple base
distribution to data \citep{lipman2023flow,liu2023flow}. Its progress in image
generation has come from a broad design space, including noise schedules and
preconditioning \citep{karras2022edm}, transformer backbones
\citep{peebles2023dit,ma2024sit,esser2024sd3}, guidance mechanisms
\citep{dhariwal2021adm,ho2022cfg,hu2023sgdm,karras2024ag}, inference-time
scaling \citep{ma2025its}, and latent representations, from latent diffusion
autoencoders to RAEs \citep{rombach2022ldm,tong2026rae,singh2026raev2}.
These advances raise a practical question:
beyond changing the sampler, architecture, or training objective, can a frozen
flow-matching generator benefit from additional computation inside its denoiser?

In parallel, looped transformer computation has re-emerged as a way to spend
extra computation through repeated hidden-state updates. Universal and looped
transformers study recurrence as a mechanism for iterative refinement, algorithm
learning, and latent reasoning
\citep{dehghani2019ut,giannou2023pc,yang2024lt,saunshi2025lt}. More recent
training-free looped transformers suggest that pretrained pre-norm transformers
can sometimes benefit from repeated layer evaluations without updating weights
\citep{chen2026looped}. However, it remains unclear how such hidden-state
looping should be used inside flow-matching image generators, where denoising
time, layer depth, token routing, and guidance all interact.

Our training-free inference-time approach repeatedly applies selected
transformer layers inside a frozen flow-matching denoiser, while leaving the
model weights, text conditioning, latent autoencoder, and outer sampler
unchanged. We organize the design around four questions: what is recomputed,
when looping is active, where it is applied, and how its effect is controlled.
Dense Token Loop repeats transformer-layer computation for all tokens, whereas
Sparse Token Loop limits repetition to a subset of tokens to reduce
computation. Sampling-Progress Gating specifies the active interval along
sampling progress, and the loop layer range specifies where looping occurs.
Loop count and strength control the repeated updates, while Loop Guidance
determines how ordinary and looped vector-field predictions are combined.

Figure~\ref{fig:qualitative_main} presents prompt-matched qualitative
comparisons and quality--efficiency trade-offs. The examples suggest that our
approach primarily corrects subject-level structure: malformed or incomplete
shapes become more coherent, spurious parts are suppressed, and object
identities, attributes, and simple counts are represented more faithfully. The
efficiency plots further show that internal layer looping reaches higher
GenEval and DPG-Bench scores than increasing only the number of outer sampling
steps, at comparable or lower generation time.

Our contributions are:
\begin{itemize}
\item We formulate internal transformer-layer looping as a training-free
mechanism for allocating additional inference compute within frozen
flow-matching image generators.
\item We organize internal loop design by what, when, and where computation is
repeated: Dense and Sparse Token Loop operate on all or selected tokens,
Sampling-Progress Gating sets the active sampling interval, and the loop layer
range sets the repeated layers.
\item We introduce Loop Guidance as an orthogonal prediction-space mechanism
that controls how the loop-induced correction is applied by combining ordinary
and looped vector-field predictions.
\item We evaluate and analyze the loop design on Scale-RAE and RAEv2 through
ablations of loop count, layer range, sampling-progress interval, sparsifier
design, and guidance.
\end{itemize}

% ----- Inlined from 1_related.tex -----
\section{Related Work}

\paragraph{Flow matching and diffusion-transformer design.}
Flow matching provides a flexible generative modeling framework by learning
vector fields between base and data distributions
\citep{lipman2023flow,liu2023flow}. In image generation, this formulation has
been combined with diffusion-transformer backbones
\citep{peebles2023dit,ma2024sit,esser2024sd3}, improved schedule and preconditioning
choices \citep{karras2022edm}, and increasingly strong latent representations
\citep{rombach2022ldm,tong2026rae,singh2026raev2}. Prior work also studies
latent-space manipulation in transformer-based flow matching \citep{hu2024lfm}.
Our work does not introduce a new training objective, backbone, sampler, or
autoencoder. Instead, it studies whether a frozen flow-matching transformer can
use additional inference compute through internal hidden-state re-evaluation.

\paragraph{Inference-time control and guidance.}
A large body of work improves diffusion and flow models at inference time
through schedule design, guidance, candidate selection, or prediction-space
modification. Classifier guidance, classifier-free guidance, and self-guidance
alter the denoising prediction without changing the generator weights
\citep{dhariwal2021adm,ho2022cfg,hu2023sgdm}, while autoguidance combines
predictions from models of different quality \citep{karras2024ag}. Other
approaches scale
inference by adding denoising steps, searching over noisy candidates, or
applying rejection-style selection \citep{ma2025its,na2024diffrs}. Token
routing and token merging have also been used to improve diffusion training or
inference efficiency \citep{krause2025tread,bolya2023tomesd}; our Sparse Token
Loop instead applies token selection only at inference inside a frozen denoiser
while retaining full-token context. Loop Guidance is related in spirit
to prediction-space control, but its direction is defined by two paths through
the same frozen model: an ordinary path and a looped path.

\begin{figure*}[t]
    \centering
    \includegraphics[width=0.94\textwidth]{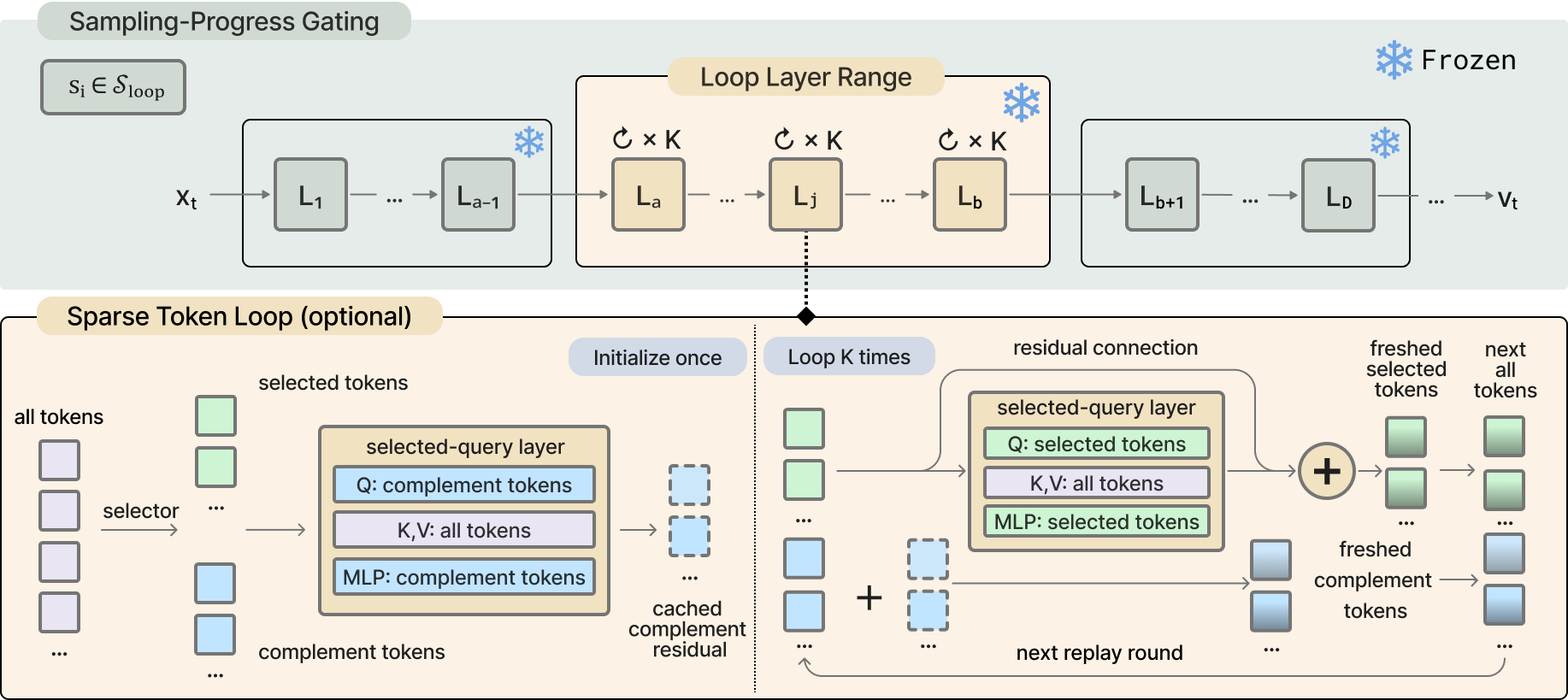}
    \caption{Overview of training-free internal looping. Sampling-Progress
    Gating determines when looping is active. Dense Token Loop repeats the
    selected layer range for all tokens, while the optional Sparse Token Loop
    restricts repetition to selected tokens and reuses a complement residual
    initialized once. All weights remain frozen, and the outer sampler is
    unchanged.}
    \label{fig:method_overview}
\end{figure*}

\paragraph{Looped transformers.}
Recurrent transformer computation has been studied in universal transformers
and recent looped-transformer models, where repeated application of shared
layers can support iterative computation, algorithm learning, or latent
reasoning \citep{dehghani2019ut,giannou2023pc,yang2024lt,saunshi2025lt}.
Training-free looped transformers further show that repeated layer evaluation
can be useful even without retraining \citep{chen2026looped}.
Our method differs by applying looped computation inside frozen flow-matching
image denoisers and by separating Dense Token Loop, Sampling-Progress Gating,
Sparse Token Loop with cached complement reuse, and prediction-space Loop
Guidance.

% ----- Inlined from 2_method.tex -----
\section{Method}

\subsection{Preliminaries}
\label{sec:method_preliminaries}

Flow matching learns a time-dependent vector field
$\mathbf{v}_\theta(\mathbf{x}_t,t,c)$ whose numerical integration transports
an initial noise state toward the data distribution
\citep{lipman2023flow,liu2023flow}; here $\mathbf{x}_t$ is the latent image
state, $t$ is flow time, and $c$ is the text condition. In DiT-based image
generators, this vector field is predicted by residual transformer layers
conditioned on flow time and text \citep{peebles2023dit,esser2024sd3}. Each
denoising call ordinarily evaluates every transformer layer once. We index the
denoiser's $D$ shape-preserving transformer layers by
$j\in\{0,\ldots,D-1\}$ and write $\mathbf{L}_j$ for layer $j$. We use
$\mathbf{h}_j$ for the full token hidden state entering $\mathbf{L}_j$.

\paragraph{Naive Internal Looping.}
A direct way to add inference compute is to reapply the same transformer layer,
or layer range, $K$ times at every denoising call. For one layer, the naive
update can be written as
\begin{equation}
\begin{aligned}
\mathbf{Z}^{(0)}
&=\mathbf{h}_j,\\
\mathbf{Z}^{(r)}
&=\mathbf{L}_j(\mathbf{Z}^{(r-1)};t,c),
\quad r=1,\ldots,K,\\
\mathbf{h}_{j+1}
&=\mathbf{Z}^{(K)}.
\end{aligned}
\label{eq:naive_internal_loop}
\end{equation}
This provides additional hidden-state refinement at fixed
$(\mathbf{x}_t,t,c)$, before the vector-field prediction is returned. However,
directly repeating full layer updates can produce an excessively large
hidden-state displacement, applying the same amount of replay everywhere ignores
the different roles of layer depth and sampling progress, and recomputing every
token introduces substantial cost. These limitations motivate our controlled
formulation, which normalizes each repeated residual and determines where, when,
and for which tokens looping is applied.

\subsection{Our Method}

We add inference compute inside a frozen flow-matching denoiser by repeatedly
applying a selected transformer-layer range during chosen denoising calls, an
operation we call a loop.
Figure~\ref{fig:method_overview} organizes the design around four choices.
\emph{What} is repeated is set by token scope: Dense Token Loop covers all
tokens, while the optional Sparse Token Loop restricts repetition to selected
tokens and reuses a complement residual initialized once.
Sampling-Progress Gating determines \emph{when} looping is active, and the loop
layer range determines \emph{where}. Loop count and strength control repeated
updates, while Loop Guidance determines \emph{how} ordinary and looped
vector-field predictions are combined. Calls outside the active interval and
layers outside the selected range follow the ordinary path.

\subsection{Dense Token Loop}
\label{sec:full_token_method}

Dense Token Loop keeps the repeated-layer structure in
Eq.~\eqref{eq:naive_internal_loop} but controls each round as a residual update.
We define the residual map of layer $j$ together with the ordinary update:
\begin{equation}
\begin{aligned}
\mathbf{R}_j(\mathbf{Z};t,c)
&=\mathbf{L}_j(\mathbf{Z};t,c)-\mathbf{Z},\\
\mathbf{h}_{j+1}
&=\mathbf{h}_j+\mathbf{R}_j(\mathbf{h}_j;t,c)
=\mathbf{L}_j(\mathbf{h}_j;t,c).
\end{aligned}
\label{eq:layer_residual}
\end{equation}

It then repeats a selected layer for $r=1,\ldots,K$ rounds with per-round
strength $\eta=\lambda_{\mathrm{loop}}/K$:
\begin{equation}
\begin{aligned}
\mathbf{Z}^{(0)}
&=\mathbf{h}_j,\\
\mathbf{Z}^{(r)}
&=\mathbf{Z}^{(r-1)}
  +\eta\mathbf{R}_j(\mathbf{Z}^{(r-1)};t,c),\\
\mathbf{h}_{j+1}
&=\mathbf{Z}^{(K)}.
\end{aligned}
\label{eq:loop_update}
\end{equation}
Here $K$ denotes the loop count, equal to the number of evaluations of the
frozen layer, and $\lambda_{\mathrm{loop}}$ denotes the total loop strength
distributed across the $K$ rounds. We set $\lambda_{\mathrm{loop}}=1$ in all
subsequent experiments and omit it hereafter, so $\eta=1/K$. With $K=1$,
Eq.~\eqref{eq:loop_update} recovers the ordinary layer output.

Let $[a,b]$ denote the selected layer range
$\{a,a+1,\ldots,b\}$. Its ordinary map is
$\mathbf{G}_{a:b}=\mathbf{L}_b\circ\cdots\circ\mathbf{L}_a$. We apply
Eq.~\eqref{eq:loop_update} successively from $\mathbf{L}_a$ through
$\mathbf{L}_b$ and denote the resulting composition by
$\boldsymbol{\Phi}_{a:b}$.

\subsection{Sampling-Progress Gating}
\label{sec:timestep_loop_method}

Denoising follows a coarse-to-fine trajectory: high-noise calls strongly
influence text-aligned content and global layout, whereas low-noise calls
mainly remove residual noise and sharpen local details
\citep{balaji2022ediffi,park2024dtr,wang2023painters}. We therefore anchor the
loop window at the noise side and vary only its endpoint.

For $N$ denoising calls, let $t_i$ be the flow time and
$s_i=i/(N-1)$ the normalized sampling progress, where
$i=0,\ldots,N-1$, so $s_i=0$ is the initial noise state and $s_i=1$
is the data endpoint. Given $s_{\mathrm{end}}\in[0,1]$, we define the
loop-active interval as
$\mathcal{S}_{\mathrm{loop}}=[0,s_{\mathrm{end}}]$.
Sampling-Progress Gating applies
\begin{equation}
\mathbf{h}_{b+1} =
\begin{cases}
\boldsymbol{\Phi}_{a:b}(\mathbf{h}_a;t_i,c),
& s_i\in\mathcal{S}_{\mathrm{loop}},\\
\mathbf{G}_{a:b}(\mathbf{h}_a;t_i,c),
& \text{otherwise}.
\end{cases}
\label{eq:timestep_gate}
\end{equation}
Calls with $s_i>s_{\mathrm{end}}$ follow the ordinary denoiser path.

\subsection{Sparse Token Loop}
\label{sec:token_method}

Sparse Token Loop reduces repeated token computation by looping over a token
subset. Let $U$ denote the tokens eligible for sparse replay under the chosen
sparsifier, and let $\rho_{\mathrm{sel}}\in(0,1]$ denote the fraction selected
for repeated computation. At each active denoising call and selected layer
$\mathbf{L}_j$, we uniformly sample $X\subseteq U$, with
$|X|=\lceil\rho_{\mathrm{sel}}|U|\rceil$, and define
$\bar X=U\setminus X$. This partition is fixed across the $K$ rounds for that
layer. For any token subset $A$, let
$\mathbf{R}_{j,A}(\mathbf{Z};t,c)$ denote the layer residual computed only for
$A$ using the full token sequence as context. We write
$\mathbf{Z}_X^{(0)}$ and $\mathbf{Z}_{\bar X}^{(0)}$ for the corresponding
token states obtained by partitioning $\mathbf{Z}^{(0)}$, and
$\operatorname{Merge}$ restores their original token order.

\paragraph{Initialize once.}
Before looping, we compute and cache only the complement-token residual:
\begin{equation}
\mathbf{C}_{\bar X}
=\mathbf{R}_{j,\bar X}(\mathbf{Z}^{(0)};t,c).
\label{eq:sparse_token_init}
\end{equation}

\paragraph{Loop $K$ times.}
At round $r=1,\ldots,K$, we evaluate the selected-token residual on the
current all-token state and reuse $\mathbf{C}_{\bar X}$ for the complement:
\begin{equation}
\begin{aligned}
\mathbf{Z}_X^{(r)}
&=\mathbf{Z}_X^{(r-1)}
  +\eta\mathbf{R}_{j,X}(\mathbf{Z}^{(r-1)};t,c),\\
\mathbf{Z}_{\bar X}^{(r)}
&=\mathbf{Z}_{\bar X}^{(r-1)}
  +\eta\mathbf{C}_{\bar X},\\
\mathbf{Z}^{(r)}
&=\operatorname{Merge}\!\left(
  \mathbf{Z}_X^{(r)},\mathbf{Z}_{\bar X}^{(r)}
  \right).
\end{aligned}
\label{eq:sparse_token_update}
\end{equation}

This design has two advantages. First, it reduces repeated computation because
only the selected-token residual $\mathbf{R}_{j,X}$ is freshly evaluated after
the complement residual is initialized once. Second, both token groups are
updated in every round, which keeps them synchronized in loop depth.
Consequently, each selected-token attention update uses queries from $X$ and
keys and values from both $X$ and $\bar X$, whose states have undergone the
same number of loop updates, rather than using complement states frozen at
initialization. Concretely, Sparse Token Loop avoids repeated query,
attention-output, and MLP evaluation for the complement tokens while retaining
keys and values from all tokens to preserve global context. Its implementation
incurs token-partitioning, routing, and merging overhead, which partially
offsets these computational savings.
This is related to training-free feature caching for diffusion transformers, but
our cache is an intra-layer complement residual tied to the current token
partition rather than a cross-timestep feature cache \citep{zou2025toca}.

\subsection{Loop Guidance}
\label{sec:loop_guidance_method}

The preceding controls alter hidden-state computation inside one denoiser path.
Motivated by prediction-space extrapolation in classifier-free guidance and
autoguidance \citep{ho2022cfg,karras2024ag}, Loop Guidance combines an ordinary
prediction with a looped prediction produced by either Dense Token Loop or
Sparse Token Loop:
\begin{equation}
\begin{aligned}
\mathbf{v}_{\mathrm{base}}
&=\mathbf{v}_\theta(\mathbf{x}_t,t,c),\\
\mathbf{v}_{\mathrm{loop}}
&=\mathbf{v}_\theta^{\mathrm{loop}}(\mathbf{x}_t,t,c),\\
\mathbf{v}_{\mathrm{lg}}
&=\mathbf{v}_{\mathrm{base}}
  +g_{\mathrm{lg}}\!\left(
    \mathbf{v}_{\mathrm{loop}}-\mathbf{v}_{\mathrm{base}}
  \right).
\end{aligned}
\label{eq:loop_guidance}
\end{equation}
The scale $g_{\mathrm{lg}}$ controls the strength of the loop-induced
correction in prediction space, with $g_{\mathrm{lg}}>1$ extrapolating beyond
the looped prediction. Because both predictions are required, Loop Guidance
costs more than a single looped path.

% ----- Inlined from 3_exp.tex -----
\section{Experiments}

\begin{table*}[!t]
  \centering
  \begingroup
  \small
  \renewcommand{\arraystretch}{1.0}
  \setlength{\tabcolsep}{1.5pt}
  \begin{tabular}{cl|ccccccc}
  \toprule
  \textbf{Model} & \textbf{Method} & \textbf{GenEval$\uparrow$} &
  \textbf{DPG$\uparrow$} &
  \textbf{ImageReward$\uparrow$} &
  \textbf{CLIP$\uparrow$} &
  \textbf{OpenCLIP$\uparrow$} &
  \textbf{PickScore$\uparrow$} &
  \textbf{Time (s)$\downarrow$} \\
  \midrule
  & w/o Loop & 0.4471 & 0.7656 & 0.0994 &
  0.3204 & 0.3490 & 0.2122 & \textbf{1.3054} \\
  \smash{\shortstack[c]{\textit{Scale-RAE}\\\textit{DiT2.4B}}} &
  Dense Token Loop (Ours) & \textbf{0.5422} & \textbf{0.8007} & \textbf{0.4816} & \textbf{0.3302} & \textbf{0.3622} & \textbf{0.2165} & 2.1812 \\
  & Sparse Token Loop (Ours) & \underline{0.5082} & \underline{0.7893} & \underline{0.2798} & \underline{0.3249} & \underline{0.3555} & \underline{0.2144} & \underline{1.6884} \\
  \midrule\midrule
  & w/o Loop & 0.5321 & 0.8003 & 0.4609 &
  0.3285 & 0.3604 & 0.2175 & \textbf{1.6853} \\
  \smash{\shortstack[c]{\textit{Scale-RAE}\\\textit{DiT9.8B}}} &
  Dense Token Loop (Ours) & \textbf{0.5977} & \underline{0.8178} & \textbf{0.7211} & \textbf{0.3344} & \textbf{0.3682} & \textbf{0.2221} & 4.0384 \\
  & Sparse Token Loop (Ours) & \underline{0.5857} & \textbf{0.8206} & \underline{0.6365} & \underline{0.3328} & \underline{0.3663} & \underline{0.2205} & \underline{2.4410} \\
  \midrule\midrule
  % DiffusionBench reports this SigLIP2-B DDT as 615M parameters; direct
  % counting of the checkpoint model state gives 615,411,540 parameters.
  & w/o Loop & \textbf{0.3829} & 0.7131 &
  -0.3385 & 0.3076 & 0.3176 & 0.2055 & \textbf{0.8025} \\
  \smash{\shortstack[c]{\textit{RAEv2}\\\textit{SigLIP2-B (615M)}}} &
  Dense Token Loop (Ours) & 0.3795 & \textbf{0.7226} & \textbf{-0.2538} & \textbf{0.3101} & \textbf{0.3202} & \textbf{0.2062} & \underline{1.1023} \\
  & Sparse Token Loop (Ours) & \underline{0.3815} & \underline{0.7152} & \underline{-0.2931} & \underline{0.3086} & \underline{0.3187} & \underline{0.2059} & 1.3572 \\
  \bottomrule
  \end{tabular}
  \endgroup
  \caption{Main results across three RAE-based model configurations. w/o Loop
  denotes ordinary denoising. The best and second-best values for each model
  are shown in bold and underlined, respectively.}
  \label{tab:method_overview}
\end{table*}

\begin{table*}[!t]
\centering
\begingroup
\small
\renewcommand{\arraystretch}{1.0}
\setlength{\tabcolsep}{1.5pt}
\begin{tabular}{clc|cccccc}
\toprule
\textbf{Model} & \textbf{Method} & \textbf{$\mathbf{g}_{\mathrm{lg}}$} &
\textbf{GenEval$\uparrow$} & \textbf{DPG$\uparrow$} &
\textbf{ImageReward$\uparrow$} &
\textbf{CLIP$\uparrow$} &
\textbf{OpenCLIP$\uparrow$} &
\textbf{PickScore$\uparrow$} \\
\midrule
\smash{\raisebox{-10pt}{\shortstack[c]{
\textit{Scale-RAE}\\
\textit{DiT2.4B}
}}} &
w/o Loop & 0.0 & 0.4471 & 0.7656 & 0.0994 & 0.3204 & 0.3490 & 0.2122 \\
& Dense Token Loop (Ours) & 2.8 & \textbf{0.5691} & \textbf{0.8053} & \textbf{0.5622} & \textbf{0.3306} & \textbf{0.3627} & \textbf{0.2171} \\
\midrule\midrule
\smash{\raisebox{-10pt}{\shortstack[c]{
\textit{Scale-RAE}\\
\textit{DiT9.8B}
}}} &
w/o Loop & 0.0 & 0.5321 & 0.8003 & 0.4609 & 0.3285 & 0.3604 & 0.2175 \\
& Dense Token Loop (Ours) & 6.0 & \textbf{0.6432} & \textbf{0.8264} & \textbf{0.7975} & \textbf{0.3332} & \textbf{0.3661} & \textbf{0.2229} \\
\midrule\midrule
& w/o Loop & 0.0 & 0.3829 & 0.7131 & -0.3385 & 0.3076 & 0.3176 & 0.2055 \\
\smash{\raisebox{-5pt}{\shortstack[c]{
\textit{RAEv2}\\
\textit{SigLIP2-B (615M)}
}}} &
Dense Token Loop (Ours) & 9.0 & \textbf{0.4532} & \textbf{0.7521} & \textbf{0.0460} & \textbf{0.3159} & \textbf{0.3297} & \textbf{0.2090} \\
& Sparse Token Loop (Ours) & 6.0 & 0.4114 & 0.7429 & -0.1296 & 0.3129 & 0.3251 & 0.2074 \\
\bottomrule
\end{tabular}
\endgroup
\caption{Loop Guidance improves both primary metrics and all auxiliary metrics
across the three evaluated generators. Baseline rows use $g_{\mathrm{lg}}=0$,
and guided rows use the listed scales. Best values within each model are
bolded.}
\label{tab:loop_guidance_results}
\end{table*}

\subsection{Experimental Setup}

We conduct experiments on two Scale-RAE configurations,
Qwen1.5B/DiT2.4B and Qwen7B/DiT9.8B, and on RAEv2.
For RAEv2, we use the public DiffusionBench RAE-SigLIP2-B checkpoint with a
615M-parameter DDT~\cite{leng2026diffusionbench}.
All pretrained components remain frozen throughout.
We identify the shared loop configuration on Scale-RAE Qwen1.5B/DiT2.4B and
directly transfer it to Qwen7B/DiT9.8B and RAE-SigLIP2-B; the transferred
setting remains effective on both target models.
We use GenEval~\citep{ghosh2023geneval} and
DPG-Bench~\citep{hu2024ella} as the primary metrics, together with
CLIP Score~\citep{radford2021clip},
OpenCLIP Score~\citep{cherti2023openclip},
PickScore~\citep{kirstain2023pick}, and
ImageReward~\citep{xu2023imagereward}.
Detailed checkpoints and evaluation protocols are provided in the appendix.

Controlled ablations use Scale-RAE Qwen1.5B/DiT2.4B with 25 Euler steps,
$K=4$, loop layer range $[12,27]$, and $s_i\in[0,0.5]$; each ablation changes
only the named factor. RAE-SigLIP2-B retains its 50-step Euler schedule.
We use 25 outer sampling steps because increasing the schedule to 50 steps
substantially increases inference time while yielding only marginal
improvements in GenEval and DPG-Bench. Complete configurations are provided
in the appendix.
\subsection{Main Results}

Table~\ref{tab:method_overview} compares the unguided loop variants across
three frozen generators. Both Dense Token Loop and Sparse Token Loop
improve GenEval and DPG-Bench at both Scale-RAE scales. Overall, Dense Token
Loop achieves stronger quality gains but requires more generation time,
whereas Sparse Token Loop is slightly weaker but more efficient overall. On
RAEv2, both variants improve DPG-Bench and all auxiliary metrics but slightly
reduce GenEval, so the unguided transfer result is mixed. We therefore avoid
claiming a uniformly positive RAEv2 gain from the transferred Scale-RAE
hyperparameters. As shown in Table~\ref{tab:loop_guidance_results}, Loop
Guidance produces the clearer RAEv2 improvement.

\begin{figure*}[!t]
  \centering
  \includegraphics[width=0.94\textwidth]{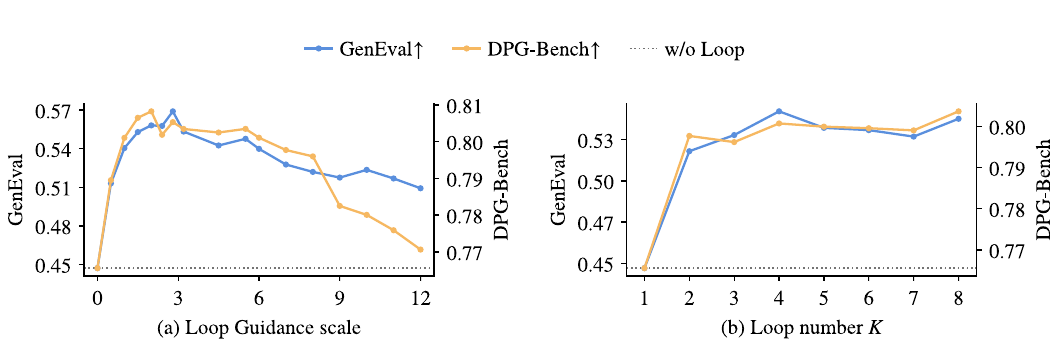}
  \caption{Moderate Loop Guidance scales and loop counts capture most quality
  gains on Scale-RAE (DiT2.4B). Panel (a) varies the Loop Guidance scale, and
  panel (b) varies the loop count $K$.}
  \label{fig:guidance_main}
  \label{fig:k_main}
  \end{figure*}

\paragraph{Sparse Token Loop.}
As described in the Sparse Token Loop subsection, the method reduces repeated
complement-token computation while preserving full-token attention context. On
both Scale-RAE models, it remains above w/o Loop on all reported
quality metrics while reducing generation time relative to Dense Token Loop.
The reduction is substantially larger on DiT9.8B than on DiT2.4B. On the
smaller 615M DDT, however, Sparse Token Loop is slower than Dense Token Loop,
suggesting that its additional routing overhead is not offset at this scale.
Specifically, Sparse Token Loop reduces generation time by 22.6\% on DiT2.4B
and 39.6\% on DiT9.8B relative to Dense Token Loop.

\paragraph{Loop Guidance.}
We do not exhaustively tune the guidance scale; for each model,
Table~\ref{tab:loop_guidance_results} reports one setting that performs
reasonably well in our trials. Combining the ordinary and looped predictions
improves both primary metrics across the Scale-RAE DiT2.4B and DiT9.8B
checkpoints and RAEv2, together with consistent gains on the
auxiliary metrics. In particular, although the unguided improvements on RAEv2
are small, adding Loop Guidance produces clear improvements on both primary
metrics. On RAEv2, Loop Guidance raises GenEval from 0.3829 to 0.4532 and
DPG-Bench from 0.7131 to 0.7521 while improving every auxiliary metric.
This suggests that the vector-field difference introduced by looping,
$\mathbf{v}_{\mathrm{loop}}-\mathbf{v}_{\mathrm{base}}$, encodes a small,
directionally useful quality correction that Loop Guidance can amplify.

\subsubsection{Quality--Efficiency Tradeoff}

As shown in the right column of Figure~\ref{fig:qualitative_main}, internal
layer looping uses inference compute more effectively than additional outer
sampling steps. A 25-step Dense Token Loop reaches 0.5422 GenEval and 0.8007
DPG-Bench in 2.1812\,s, outperforming the 50-step no-loop baseline
(0.4742 and 0.7738) while remaining faster than its 2.4032\,s runtime.
Sparse Token Loop similarly reaches 0.5082 GenEval and 0.7893 DPG-Bench in
1.6883\,s, compared with 0.4770 and 0.7673 in 1.7575\,s for the 35-step
no-loop baseline. These comparisons indicate that the gains arise from
allocating computation to internal refinement rather than merely increasing
total inference cost. The latency advantage remains despite repeated
evaluations of the selected layer range, showing that internal compute can be
allocated more selectively than full-denoiser resampling.

\FloatBarrier

  % Aggregate scores and evaluator bindings were audited before release.

  \begin{figure*}[!t]
  \centering
  \includegraphics[width=0.88\textwidth]{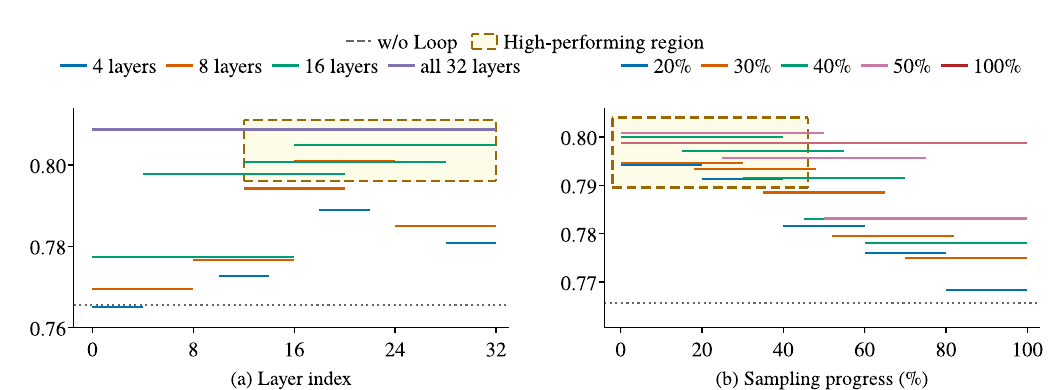}
  \caption{Mid-to-late layer ranges and earlier loop-active sampling intervals
  give the strongest DPG-Bench settings on Scale-RAE (DiT2.4B). Panel (a)
  varies the loop layer range with the loop-active sampling-progress interval
  fixed; panel (b) varies the loop-active sampling-progress interval with the
  layer range fixed. The corresponding GenEval panels are provided in the
  appendix.}
  \label{fig:window_ablation_main}
  \end{figure*}

  \subsection{Loop Configuration Ablations}

  Using Scale-RAE DiT2.4B with 25 outer sampling steps, we ablate the Loop
  Guidance scale, loop count, loop layer range, and loop-active
  sampling-progress interval. Unless the named factor is varied, we use a loop
  count of $K=4$, loop layer range $[12,27]$, and loop-active
  sampling-progress interval $s_i\in[0,0.5]$.

  \paragraph{Loop Guidance Scale.}
  With the loop configuration fixed, Figure~\ref{fig:guidance_main}(a) varies
  $g_{\mathrm{lg}}$. Moderate settings improve both primary metrics, although
  their preferred scales differ, while overly large scales weaken the gains.
  This supports moderate prediction-space extrapolation.

  \paragraph{Loop Count.}
  With the layer range and sampling-progress interval fixed,
  Figure~\ref{fig:k_main}(b) varies the loop count. Moderate loop
  counts capture most of the gains, whereas additional loop rounds give
  diminishing, metric-dependent returns while increasing computation. We
  therefore use a loop count of $K=4$ by default.
  % Aggregate scores and evaluator bindings were audited before release.

  \paragraph{Loop Layer Range.}
  With a loop count of $K=4$ and the sampling-progress interval fixed,
  Figure~\ref{fig:window_ablation_main}(a) compares different layer
  windows on DPG-Bench.
  Middle-to-late layer ranges consistently outperform early ranges. Looping
  over all 32 layers provides little additional benefit over a well-placed
  16-layer range while requiring more computation. We therefore use
  $[12,27]$ as the default loop layer range.

  % The released configuration and aggregate data reconstruct this sweep.

  \paragraph{Loop-Active Sampling-Progress Interval.}
  With a loop count of $K=4$ and the layer range fixed,
  Figure~\ref{fig:window_ablation_main}(b) varies when the loop is active along
  the sampling trajectory on DPG-Bench. Across interval lengths, looping during
  earlier, higher-noise stages is generally more effective than looping later.
  Table~\ref{tab:sampling_progress_gate} isolates the no-loop, full-schedule,
  and default gated policies. Activating the loop over the first half of
  sampling, $s_i\in[0,0.5]$, retains most of the full-schedule benefit with
  less loop computation and is therefore our default.

  \begin{table}[t]
  \centering
  \begingroup
  \small
  \setlength{\tabcolsep}{5pt}
  \renewcommand{\arraystretch}{1.0}
  \begin{tabular}{@{}c|ccc@{}}
  \toprule
  \textbf{Loop-active interval} & \textbf{GenEval$\uparrow$} &
  \textbf{DPG$\uparrow$} &
  \textbf{Time(s)$\downarrow$} \\
  \midrule
  w/o Loop & 0.4471 & 0.7656 & \text{1.3054} \\
  $[0,1]$ & \text{0.5597} & 0.7987 & 2.9957 \\
  $[0,0.5]$ & 0.5422 & \text{0.8007} & 2.1812 \\
  \bottomrule
  \end{tabular}
  \endgroup
  \caption{For Dense Token Loop, $[0,1]$ denotes full-schedule looping and
  $[0,0.5]$ denotes Sampling-Progress Gating; the latter preserves most
  full-schedule Loop quality while reducing generation time on Scale-RAE.}
  \label{tab:sampling_progress_gate}
  \end{table}

  \subsection{Further Exploration}

  \paragraph{Sparsifier Comparison.}
  Since Sparse Token Loop concentrates repeated computation on selected
  tokens, a natural question is whether prioritizing tokens that are important
  for refinement can improve generation quality. On Scale-RAE,
  Table~\ref{tab:sparsifier_main} compares random routing with condition-aware
  and attention-aware routing, which prioritize tokens that respond more
  strongly to conditioning and attention, respectively. Neither targeted
  variant clearly improves upon random. On RAEv2, whose DDT operates on a joint
  sequence of image and conditioning tokens, we additionally compare random
  routing with an image/condition split; their results remain close. Overall,
  the specific sparsifier designs are described in the appendix. More
  effective sparsifiers may exist, but random routing gives the strongest
  overall balance among the choices tested here.

  % The released aggregate data reconstruct this comparison.
  \begin{table}[t]
  \centering
  \begingroup
  \small
  \renewcommand{\arraystretch}{1.0}
  \setlength{\tabcolsep}{3pt}
  \begin{tabular}{clc|cc}
  \toprule
  \textbf{Model} & \textbf{Sparsifier} & \textbf{Ratio} &
  \textbf{GenEval$\uparrow$} &
  \textbf{DPG-Bench$\uparrow$} \\
  \midrule
  & w/o Loop & -- & 0.4471 & 0.7656 \\
  & random & 50\% & 0.5029 & 0.7951 \\
  & condition-aware & 50\% & 0.5039 & 0.7846 \\
  \smash{\textit{Scale-RAE}} &
  attention-aware & 50\% & 0.5008 & 0.7894 \\
  & random & 70\% & 0.5140 & 0.7962 \\
  & condition-aware & 70\% & 0.5175 & 0.7918 \\
  & attention-aware & 70\% & 0.5070 & 0.7918 \\
  \midrule\midrule
  & w/o Loop & -- & 0.3829 & 0.7131 \\
  \smash{\textit{RAEv2}} &
  random & 50\% & 0.4114 & 0.7429 \\
  & image/condition & -- & 0.4170 & 0.7474 \\
  \bottomrule
  \end{tabular}
  \endgroup
  \caption{Sparsifier comparison on Scale-RAE and RAEv2.}
  \label{tab:sparsifier_main}
  \end{table}

  \paragraph{VAE-Based Generators.}
  Table~\ref{tab:vae_transfer} reports a transfer diagnostic on
  PixArt-alpha~\citep{chen2024pixartalpha} and
  FLUX.2~\citep{bfl2025flux2}. For both generators, we apply Loop Guidance
  with $g_{\mathrm{lg}}=3.0$. Neither model shows a consistent improvement
  across both primary metrics. We therefore hypothesize that the benefit
  observed on RAE-based generators may not transfer directly to VAE-based
  generators; understanding the underlying reason requires further investigation.

  \begin{table}[t]
  \centering
  \begingroup
  \small
  \renewcommand{\arraystretch}{1.0}
  \setlength{\tabcolsep}{4pt}
  \begin{tabular}{cl|cc}
  \toprule
  \textbf{Model} & \textbf{Method} & \textbf{GenEval$\uparrow$} &
  \textbf{DPG-Bench$\uparrow$} \\
  \midrule
  & w/o Loop & 0.4778 & 0.7038 \\
  \textit{PixArt-alpha} & Dense Token Loop & 0.4710 & 0.7170 \\
  & Loop Guidance & 0.3928 & 0.6986 \\
  \midrule\midrule
  & w/o Loop & 0.8459 & 0.8740 \\
  \textit{FLUX.2} & Dense Token Loop & 0.8482 & 0.8740 \\
  & Loop Guidance & 0.8310 & 0.8720 \\
  \bottomrule
  \end{tabular}
  \endgroup
  \caption{VAE transfer diagnostic on PixArt-alpha and FLUX.2.}
  \label{tab:vae_transfer}
  \end{table}

  % Aggregate scores and evaluator bindings were audited before release.
  \paragraph{Qualitative Correction Effect of Looping.}
  The left column of Figure~\ref{fig:qualitative_main} offers a preliminary
  view of what looping corrects under matched prompts. Corrections are
  mainly subject-level: malformed or incomplete shapes become coherent,
  spurious parts are suppressed, and attributes bind more cleanly. The shown
  GenEval cases also have clearer identities and simple counts, while Loop
  Guidance gives cleaner DPG-Bench presentations. Appendix GenEval2 examples
  suggest that harder counting and placement errors remain. Thus, subject-level
  correction appears stronger, but its mechanism remains open.

% ----- Inlined from 4_conclusion.tex -----
\section{Conclusion}

We show that inference computation can be added inside a frozen flow-matching
denoiser by repeatedly applying selected transformer layers, without changing
model weights, conditioning, latent representations, or the outer sampler.
Organizing this operation by token scope, sampling progress, layer range, and
prediction-space control yields Dense Token Loop, Sparse Token Loop,
Sampling-Progress Gating, and Loop Guidance. Across both Scale-RAE scales,
looping improves the primary and auxiliary quality metrics, while Sparse Token
Loop provides a lower-latency alternative to dense repetition. At comparable
generation time, internal looping also improves upon allocating computation
only to additional outer sampling steps. Loop Guidance further improves
GenEval and DPG-Bench on all three tested RAE-based generators.

The results also clarify the current scope of the method. Looping most
reliably corrects subject formation and attribute binding, but does not
consistently repair counting or spatial composition on GenEval2. Its effect
depends on layer location and sampling progress, and the tested VAE-based
generators do not show a consistent gain. Internal layer looping is therefore
a useful but model-dependent inference-time refinement axis. Future work
should develop task-aware sparsifiers and adaptive loop policies, and
determine which architectural and latent-representation properties make an
image generator receptive to repeated hidden-state updates.

\clearpage

\begin{small}
\bibliography{references}

@article{chen2026looped,
  author = {Chen, Lizhang and Li, Jonathan and Liang, Chen and Lao, Ni and Liu, Qiang},
  title = {Training-free looped transformers},
  journal = {arXiv preprint arXiv:2605.23872},
  year = {2026},
  url = {https://arxiv.org/abs/2605.23872}
}

@inproceedings{krause2025tread,
  author = {Krause, Felix and Phan, Timy and Gui, Ming and Baumann, Stefan Andreas and Hu, Vincent Tao and Ommer, Bjorn},
  title = {{TREAD}: token routing for efficient architecture-agnostic diffusion training},
  booktitle = {ICCV},
  year = {2025},
  url = {https://openaccess.thecvf.com/content/ICCV2025/html/Krause_TREAD_Token_Routing_for_Efficient_Architecture-agnostic_Diffusion_Training_ICCV_2025_paper.html}
}

@inproceedings{bolya2023tomesd,
  author = {Bolya, Daniel and Hoffman, Judy},
  title = {Token merging for fast stable diffusion},
  booktitle = {CVPR Workshops},
  year = {2023},
  url = {https://arxiv.org/abs/2303.17604}
}

@inproceedings{zou2025toca,
  author = {Zou, Chang and Liu, Xuyang and Liu, Ting and Huang, Siteng and Zhang, Linfeng},
  title = {Accelerating diffusion transformers with token-wise feature caching},
  booktitle = {ICLR},
  year = {2025},
  url = {https://arxiv.org/abs/2410.05317}
}

@inproceedings{dehghani2019ut,
  author = {Dehghani, Mostafa and Gouws, Stephan and Vinyals, Oriol and Uszkoreit, Jakob and Kaiser, Lukasz},
  title = {Universal transformers},
  booktitle = {ICLR},
  year = {2019},
  url = {https://arxiv.org/abs/1807.03819}
}

@inproceedings{giannou2023pc,
  author = {Giannou, Angeliki and Rajput, Shashank and Sohn, Jy-Yong and Lee, Kangwook and Lee, Jason D. and Papailiopoulos, Dimitris},
  title = {Looped transformers as programmable computers},
  booktitle = {ICML},
  year = {2023},
  url = {https://proceedings.mlr.press/v202/giannou23a.html}
}

@inproceedings{yang2024lt,
  author = {Yang, Liu and Lee, Kangwook and Nowak, Robert and Papailiopoulos, Dimitris},
  title = {Looped transformers are better at learning learning algorithms},
  booktitle = {ICLR},
  year = {2024},
  url = {https://arxiv.org/abs/2311.12424}
}

@inproceedings{saunshi2025lt,
  author = {Saunshi, Nikunj and Dikkala, Nishanth and Li, Zhiyuan and Kumar, Sanjiv and Reddi, Sashank J.},
  title = {Reasoning with latent thoughts: on the power of looped transformers},
  booktitle = {ICLR},
  year = {2025},
  url = {https://arxiv.org/abs/2502.17416}
}

@inproceedings{lipman2023flow,
  author = {Lipman, Yaron and Chen, Ricky T. Q. and Ben-Hamu, Heli and Nickel, Maximilian and Le, Matt},
  title = {Flow matching for generative modeling},
  booktitle = {ICLR},
  year = {2023}
}

@inproceedings{liu2023flow,
  author = {Liu, Xingchao and Gong, Chengyue and Liu, Qiang},
  title = {Flow straight and fast: learning to generate and transfer data with rectified flow},
  booktitle = {ICLR},
  year = {2023}
}

@inproceedings{hu2024lfm,
  author = {Hu, Vincent Tao and Zhang, David W. and Mettes, Pascal and Tang, Meng and Zhao, Deli and Snoek, Cees G. M.},
  title = {Latent space editing in transformer-based flow matching},
  booktitle = {AAAI},
  year = {2024},
  url = {https://arxiv.org/abs/2312.10825}
}

@inproceedings{peebles2023dit,
  author = {Peebles, William and Xie, Saining},
  title = {Scalable diffusion models with transformers},
  booktitle = {ICCV},
  year = {2023}
}

@inproceedings{ma2024sit,
  author = {Ma, Nanye and Goldstein, Mark and Albergo, Michael S. and Boffi, Nicholas M. and Vanden-Eijnden, Eric and Xie, Saining},
  title = {{SiT}: exploring flow and diffusion-based generative models with scalable interpolant transformers},
  booktitle = {ECCV},
  year = {2024},
  url = {https://arxiv.org/abs/2401.08740}
}

@inproceedings{esser2024sd3,
  author = {Esser, Patrick and Kulal, Sumith and Blattmann, Andreas and Entezari, Rahim and Muller, Jonas and Saini, Harry and Levi, Yam and Lorenz, Dominik and Sauer, Axel and Boesel, Frederic and Podell, Dustin and Dockhorn, Tim and English, Zion and Rombach, Robin},
  title = {Scaling rectified flow transformers for high-resolution image synthesis},
  booktitle = {ICML},
  year = {2024},
  url = {https://proceedings.mlr.press/v235/esser24a.html}
}

@inproceedings{rombach2022ldm,
  author = {Rombach, Robin and Blattmann, Andreas and Lorenz, Dominik and Esser, Patrick and Ommer, Bjorn},
  title = {High-resolution image synthesis with latent diffusion models},
  booktitle = {CVPR},
  year = {2022},
  url = {https://arxiv.org/abs/2112.10752}
}

@inproceedings{karras2022edm,
  author = {Karras, Tero and Aittala, Miika and Aila, Timo and Laine, Samuli},
  title = {Elucidating the design space of diffusion-based generative models},
  booktitle = {NeurIPS},
  year = {2022},
  url = {https://arxiv.org/abs/2206.00364}
}

@misc{bfl2025flux2,
  author = {{Black Forest Labs}},
  title = {{FLUX.2}: frontier visual intelligence},
  year = {2025},
  url = {https://bfl.ai/blog/flux-2}
}

@article{balaji2022ediffi,
  author = {Balaji, Yogesh and Nah, Seungjun and Huang, Xun and Vahdat, Arash and Song, Jiaming and Zhang, Qinsheng and Kreis, Karsten and Aittala, Miika and Aila, Timo and Laine, Samuli and Catanzaro, Bryan and Karras, Tero and Liu, Ming-Yu},
  title = {{eDiff-I}: text-to-image diffusion models with an ensemble of expert denoisers},
  journal = {arXiv preprint arXiv:2211.01324},
  year = {2022},
  url = {https://arxiv.org/abs/2211.01324}
}

@inproceedings{park2024dtr,
  author = {Park, Byeongjun and Woo, Sangmin and Go, Hyojun and Kim, Jin-Young and Kim, Changick},
  title = {Denoising task routing for diffusion models},
  booktitle = {ICLR},
  year = {2024},
  url = {https://proceedings.iclr.cc/paper_files/paper/2024/hash/2b00b3331bd0f5fbfdd966ac06338f6d-Abstract-Conference.html}
}

@article{wang2023painters,
  author = {Wang, Binxu and Vastola, John J.},
  title = {Diffusion models generate images like painters: an analytical theory of outline first, details later},
  journal = {arXiv preprint arXiv:2303.02490},
  year = {2023},
  url = {https://arxiv.org/abs/2303.02490}
}

@inproceedings{ghosh2023geneval,
  author = {Ghosh, Dhruba and Hajishirzi, Hannaneh and Schmidt, Ludwig},
  title = {{GenEval}: an object-focused framework for evaluating text-to-image alignment},
  booktitle = {NeurIPS},
  year = {2023},
  url = {https://openreview.net/forum?id=Wbr51vK331}
}

@inproceedings{cheng2022mask2former,
  author = {Cheng, Bowen and Misra, Ishan and Schwing, Alexander G. and Kirillov, Alexander and Girdhar, Rohit},
  title = {Masked-attention mask transformer for universal image segmentation},
  booktitle = {CVPR},
  year = {2022},
  url = {https://arxiv.org/abs/2112.01527}
}

@article{kamath2025geneval2,
  author = {Kamath, Amita and Chang, Kai-Wei and Krishna, Ranjay and Zettlemoyer, Luke and Hu, Yushi and Ghazvininejad, Marjan},
  title = {{GenEval} 2: addressing benchmark drift in text-to-image evaluation},
  journal = {arXiv preprint arXiv:2512.16853},
  year = {2025},
  url = {https://arxiv.org/abs/2512.16853}
}

@inproceedings{hu2023tifa,
  author = {Hu, Yushi and Liu, Benlin and Kasai, Jungo and Wang, Yizhong and Ostendorf, Mari and Krishna, Ranjay and Smith, Noah A.},
  title = {{TIFA}: accurate and interpretable text-to-image faithfulness evaluation with question answering},
  booktitle = {ICCV},
  year = {2023},
  url = {https://arxiv.org/abs/2303.11897}
}

@article{hu2024ella,
  author = {Hu, Xiwei and Wang, Rui and Fang, Yixiao and Fu, Bin and Cheng, Pei and Yu, Gang},
  title = {{ELLA}: equip diffusion models with {LLM} for enhanced semantic alignment},
  journal = {arXiv preprint arXiv:2403.05135},
  year = {2024},
  url = {https://arxiv.org/abs/2403.05135}
}

@inproceedings{li2022mplug,
  author = {Li, Chenliang and Xu, Haiyang and Tian, Junfeng and Wang, Wei and Yan, Ming and Bi, Bin and Ye, Jiabo and Chen, He and Xu, Guohai and Cao, Zheng and Zhang, Ji and Huang, Songfang and Huang, Fei and Zhou, Jingren and Si, Luo},
  title = {{mPLUG}: effective and efficient vision-language learning by cross-modal skip-connections},
  booktitle = {EMNLP},
  year = {2022},
  url = {https://aclanthology.org/2022.emnlp-main.488/}
}

@inproceedings{chen2024pixartalpha,
  author = {Chen, Junsong and Yu, Jincheng and Ge, Chongjian and Yao, Lewei and Xie, Enze and Wang, Zhongdao and Kwok, James and Luo, Ping and Lu, Huchuan and Li, Zhenguo},
  title = {{PixArt-$\alpha$}: fast training of diffusion transformer for photorealistic text-to-image synthesis},
  booktitle = {ICLR},
  year = {2024},
  url = {https://proceedings.iclr.cc/paper_files/paper/2024/hash/fe989bb038b5dcc44181255dd6913e43-Abstract-Conference.html}
}

@article{tong2026rae,
  author = {Tong, Shengbang and Zheng, Boyang and Wang, Ziteng and Tang, Bingda and Ma, Nanye and Brown, Ellis and Yang, Jihan and Fergus, Rob and LeCun, Yann and Xie, Saining},
  title = {Scaling text-to-image diffusion transformers with representation autoencoders},
  journal = {arXiv preprint arXiv:2601.16208},
  year = {2026},
  url = {https://arxiv.org/abs/2601.16208}
}

@article{singh2026raev2,
  author = {Singh, Jaskirat and Zheng, Boyang and Wu, Zongze and Zhang, Richard and Shechtman, Eli and Xie, Saining},
  title = {Improved baselines with representation autoencoders},
  journal = {arXiv preprint arXiv:2605.18324},
  year = {2026},
  url = {https://arxiv.org/abs/2605.18324}
}

@article{leng2026diffusionbench,
  author = {Leng, Xingjian and Singh, Jaskirat and Liang, Zhanhao and Smith, Ethan and Bell, Martin and Saha, Aninda and Yuan, Yuhui and Zheng, Liang},
  title = {{DiffusionBench}: on holistic evaluation of diffusion transformers},
  journal = {arXiv preprint arXiv:2606.24888},
  year = {2026},
  url = {https://arxiv.org/abs/2606.24888}
}

@inproceedings{dhariwal2021adm,
  author = {Dhariwal, Prafulla and Nichol, Alexander},
  title = {Diffusion models beat {GANs} on image synthesis},
  booktitle = {NeurIPS},
  year = {2021},
  url = {https://arxiv.org/abs/2105.05233}
}

@article{ho2022cfg,
  author = {Ho, Jonathan and Salimans, Tim},
  title = {Classifier-free diffusion guidance},
  journal = {arXiv preprint arXiv:2207.12598},
  year = {2022}
}

@inproceedings{hu2023sgdm,
  author = {Hu, Vincent Tao and Zhang, David W. and Asano, Yuki M. and Burghouts, Gertjan J. and Snoek, Cees G. M.},
  title = {Self-guided diffusion models},
  booktitle = {CVPR},
  year = {2023},
  url = {https://arxiv.org/abs/2210.06462}
}

@inproceedings{radford2021clip,
  author = {Radford, Alec and Kim, Jong Wook and Hallacy, Chris and Ramesh, Aditya and Goh, Gabriel and Agarwal, Sandhini and Sastry, Girish and Askell, Amanda and Mishkin, Pamela and Clark, Jack and Krueger, Gretchen and Sutskever, Ilya},
  title = {Learning transferable visual models from natural language supervision},
  booktitle = {ICML},
  year = {2021},
  url = {https://proceedings.mlr.press/v139/radford21a.html}
}

@inproceedings{cherti2023openclip,
  author = {Cherti, Mehdi and Beaumont, Romain and Wightman, Ross and Wortsman, Mitchell and Ilharco, Gabriel and Gordon, Cade and Schuhmann, Christoph and Schmidt, Ludwig and Jitsev, Jenia},
  title = {Reproducible scaling laws for contrastive language-image learning},
  booktitle = {CVPR},
  year = {2023},
  url = {https://arxiv.org/abs/2212.07143}
}

@inproceedings{xu2023imagereward,
  author = {Xu, Jiazheng and Liu, Xiao and Wu, Yuchen and Tong, Yuxuan and Li, Qinkai and Ding, Ming and Tang, Jie and Dong, Yuxiao},
  title = {{ImageReward}: learning and evaluating human preferences for text-to-image generation},
  booktitle = {NeurIPS},
  year = {2023},
  url = {https://arxiv.org/abs/2304.05977}
}

@inproceedings{kirstain2023pick,
  author = {Kirstain, Yuval and Polyak, Adam and Singer, Uriel and Matiana, Shahbuland and Penna, Joe and Levy, Omer},
  title = {Pick-a-Pic: an open dataset of user preferences for text-to-image generation},
  booktitle = {NeurIPS},
  year = {2023},
  url = {https://arxiv.org/abs/2305.01569}
}

@inproceedings{ma2025its,
  author = {Ma, Nanye and Tong, Shangyuan and Jia, Haolin and Hu, Hexiang and Su, Yu-Chuan and Zhang, Mingda and Yang, Xuan and Li, Yandong and Jaakkola, Tommi and Jia, Xuhui and Xie, Saining},
  title = {Scaling inference time compute for diffusion models},
  booktitle = {CVPR},
  year = {2025},
  url = {https://openaccess.thecvf.com/content/CVPR2025/html/Ma_Scaling_Inference_Time_Compute_for_Diffusion_Models_CVPR_2025_paper.html}
}

@inproceedings{na2024diffrs,
  author = {Na, Byeonghu and Kim, Yeongmin and Park, Minsang and Shin, Donghyeok and Kang, Wanmo and Moon, Il-Chul},
  title = {Diffusion rejection sampling},
  booktitle = {ICML},
  year = {2024},
  url = {https://proceedings.mlr.press/v235/na24a.html}
}

@inproceedings{karras2024ag,
  author = {Karras, Tero and Aittala, Miika and Kynk{\"a}{\"a}nniemi, Tuomas and Lehtinen, Jaakko and Aila, Timo and Laine, Samuli},
  title = {Guiding a diffusion model with a bad version of itself},
  booktitle = {NeurIPS},
  year = {2024},
  url = {https://arxiv.org/abs/2406.02507}
}
\end{small}

\appendix
% ----- Inlined appendix from 5_supp.tex -----
\section{Method Pseudocode}

Algorithms~\ref{alg:dense_token_loop_supp}--\ref{alg:loop_guidance_supp}
summarize the four core inference-time operations. All model parameters
remain frozen.

\begin{algorithm}[H]
\caption{Dense Token Loop for a selected layer range.}
\label{alg:dense_token_loop_supp}
\small
\begin{algorithmic}[1]
\Require Frozen layers $\{\mathbf{L}_j\}_{j=a}^b$, state $\mathbf{h}_a$,
  time $t$, condition $c$, loop count $K$, total loop strength
  $\lambda_{\mathrm{loop}}$
\Ensure Refined state $\mathbf{h}_{b+1}$
\State $\eta\gets\lambda_{\mathrm{loop}}/K$
\For{$j=a,\ldots,b$}
  \State $\mathbf{Z}^{(0)} \gets \mathbf{h}_j$
  \For{$r=1,\ldots,K$}
    \State $\mathbf{Z}^{(r)} \gets
      \mathbf{Z}^{(r-1)}+\eta[
      \mathbf{L}_j(\mathbf{Z}^{(r-1)};t,c)-\mathbf{Z}^{(r-1)}]$
  \EndFor
  \State $\mathbf{h}_{j+1} \gets \mathbf{Z}^{(K)}$
\EndFor
\State \Return $\mathbf{h}_{b+1}$
\end{algorithmic}
\end{algorithm}

\begin{algorithm}[H]
\caption{Sparse Token Loop for a selected layer range.}
\label{alg:sparse_token_loop_supp}
\small
\begin{algorithmic}[1]
\Require Frozen layers $\{\mathbf{L}_j\}_{j=a}^b$, state $\mathbf{h}_a$,
  time $t$, condition $c$, loop count $K$, total loop strength
  $\lambda_{\mathrm{loop}}$
\Require Valid token set $U$, selected-token ratio $\rho_{\mathrm{sel}}$
\Ensure Refined state $\mathbf{h}_{b+1}$
\State $\eta\gets\lambda_{\mathrm{loop}}/K$
\For{$j=a,\ldots,b$}
  \State $\mathbf{Z}^{(0)}\gets\mathbf{h}_j$;
    $m\gets\lceil\rho_{\mathrm{sel}}|U|\rceil$
  \State Sample $X\subseteq U$ uniformly with $|X|=m$;
    $\bar X\gets U\setminus X$
  \State $\mathbf{C}_{\bar X}\gets
    \mathbf{R}_{j,\bar X}(\mathbf{Z}^{(0)};t,c)$
    \Comment{initialize once}
  \For{$r=1,\ldots,K$}
    \State $\mathbf{Z}_X^{(r)}\gets
      \mathbf{Z}_X^{(r-1)}+
      \eta\mathbf{R}_{j,X}(\mathbf{Z}^{(r-1)};t,c)$
      \Comment{full context}
    \State $\mathbf{Z}_{\bar X}^{(r)}\gets
      \mathbf{Z}_{\bar X}^{(r-1)}+\eta\mathbf{C}_{\bar X}$
    \State $\mathbf{Z}^{(r)}\gets
      \operatorname{Merge}(\mathbf{Z}_X^{(r)},\mathbf{Z}_{\bar X}^{(r)})$
  \EndFor
  \State $\mathbf{h}_{j+1}\gets\mathbf{Z}^{(K)}$
\EndFor
\State \Return $\mathbf{h}_{b+1}$
\end{algorithmic}
\end{algorithm}

\begin{algorithm}[H]
\caption{Sampling-Progress Gating.}
\label{alg:sampling_progress_gating_supp}
\small
\begin{algorithmic}[1]
\Require State $\mathbf{h}_a$, time $t_i$, condition $c$, call index $i$,
  total calls $N$, layer range $[a,b]$, loop count $K$,
  total loop strength $\lambda_{\mathrm{loop}}$
\Statex \textbf{Parameter:} loop-active interval
  $\mathcal{S}_{\mathrm{loop}}$
\Ensure State $\mathbf{h}_{b+1}$
\State $s_i\gets i/(N-1)$
\If{$s_i\in\mathcal{S}_{\mathrm{loop}}$}
  \State $\mathbf{h}_{b+1}\gets
    \Call{DenseTokenLoop}{
      \{\mathbf{L}_j\}_{j=a}^b,\mathbf{h}_a,t_i,c,K,
      \lambda_{\mathrm{loop}}}$
\Else
  \State $\mathbf{h}_{b+1}\gets
    \mathbf{G}_{a:b}(\mathbf{h}_a;t_i,c)$
    \Comment{ordinary path}
\EndIf
\State \Return $\mathbf{h}_{b+1}$
\end{algorithmic}
\end{algorithm}

\begin{algorithm}[H]
\caption{Loop Guidance at one denoising call.}
\label{alg:loop_guidance_supp}
\small
\begin{algorithmic}[1]
\Require Frozen denoiser $\mathbf{v}_\theta$, input $\mathbf{x}_{t_i}$,
  time $t_i$, condition $c$, guidance scale $g_{\mathrm{lg}}$
\Require Loop configuration $([a,b],K,\lambda_{\mathrm{loop}},
  \mathcal{S}_{\mathrm{loop}})$
\Ensure Guided vector-field prediction $\mathbf{v}_{\mathrm{lg}}$
\State $\mathbf{v}_{\mathrm{base}}\gets
  \mathbf{v}_\theta(\mathbf{x}_{t_i},t_i,c)$
  \Comment{ordinary path}
\State $\mathbf{v}_{\mathrm{loop}}\gets
  \mathbf{v}_\theta^{\mathrm{gate+loop}}(\mathbf{x}_{t_i},t_i,c)$
  \Comment{Algorithms 1--3}
\State $\mathbf{v}_{\mathrm{lg}}\gets
  \mathbf{v}_{\mathrm{base}}+
  g_{\mathrm{lg}}(\mathbf{v}_{\mathrm{loop}}-\mathbf{v}_{\mathrm{base}})$
\State \Return $\mathbf{v}_{\mathrm{lg}}$
\end{algorithmic}
\end{algorithm}

\section{Experimental Details}

\subsection{Evaluation Protocol}\leavevmode\par

The standard protocol uses all 553 GenEval prompts
\citep{ghosh2023geneval}, 1065 unique DPG-Bench image prompts
\citep{hu2024ella}, and all 800 GenEval2 prompts
\citep{kamath2025geneval2}. GenEval and DPG-Bench are used for the primary
comparisons; GenEval2 is used for the additional qualitative analysis. Every
setting generates one image per prompt with seed 42.
Within each comparison, the baseline and all loop variants share the exact
prompt, initial noise seed, checkpoint, resolution, precision, sampler, outer
sampling steps, text conditioning, and autoencoder.
These complete fixed-seed benchmark comparisons remove prompt-set and
initial-noise differences between methods, but they do not estimate variation
across random seeds. We therefore do not attach confidence intervals or make
statistical-significance claims.
GenEval records are exported to the pinned \texttt{geneval-main} evaluator and
scored with its Mask2Former-based dependencies
\citep{cheng2022mask2former}. DPG-Bench records are exported to the pinned
\texttt{ELLA-main/dpg\_bench} evaluator and scored with mPLUG VQA
\citep{li2022mplug}. Scale-RAE images are evaluated at their native
$224\times224$ resolution; the DPG-Bench protocol uses one image per prompt
and a $224$-pixel crop. The RAE-SigLIP2-B proxy uses its native
$256\times256$ resolution. GenEval2 uses its official 800-prompt
soft-TIFA-style protocol.
For the four auxiliary metrics, each generated image $I_j$ is paired only
with its own generation prompt $p_j$. CLIP Score, OpenCLIP Score, and
PickScore are computed as the raw cosine similarity
\[
s_j =
\frac{f_I(I_j)^\top f_T(p_j)}
{\lVert f_I(I_j)\rVert_2\lVert f_T(p_j)\rVert_2}.
\]
The respective encoders are OpenAI CLIP ViT-B/32
\citep{radford2021clip}, OpenCLIP ViT-H/14 pretrained on LAION-2B
(s32B/b79K) \citep{cherti2023openclip}, and PickScore v1
\citep{kirstain2023pick}. The reported values do not use the
temperature-scaled logits or a batch-wise softmax. ImageReward is the raw
scalar $s_j=R(p_j,I_j)$ returned by \texttt{ImageReward-v1.0}
\citep{xu2023imagereward}. For each variant and each metric, the table reports
\[
\bar{s} =
\frac{\sum_{j\in\mathrm{GenEval}}s_j+
      \sum_{j\in\mathrm{DPG}}s_j}{553+1065},
\]
equivalently the sample-count-weighted mean of the two split means. Higher is
better for all four metrics.

\subsection{Models and Checkpoints}\leavevmode\par

We evaluate two official Scale-RAE configurations with the following exact
checkpoint identifiers \citep{tong2026rae}:
\begingroup
\raggedright
\texttt{nyu-visionx/\allowbreak Scale-RAE-\allowbreak
Qwen1.5B\_DiT2.4B}\par
\texttt{nyu-visionx/\allowbreak Scale-RAE-\allowbreak
Qwen7B\_DiT9.8B}\par
\endgroup
Both repositories retain their official FP32 parameter dtype and use model
guidance level 1.0. For architecture transfer, we load the public repository
\texttt{diffusion-bench/\allowbreak diffusion-bench}
\citep{leng2026diffusionbench} and its EMA checkpoint
\texttt{t2i-ddt-en28d1152hd72-\allowbreak
dn2d2048hd128-\allowbreak rae-siglip2-vit-b-\allowbreak vpred-t4-v1.pt}.
This 615M-parameter RAE-SigLIP2-B/DDT checkpoint is used as a proxy and is not
the unreleased official RAEv2 text-to-image checkpoint
\citep{singh2026raev2}. Its reported runs disable classifier-free guidance.
The VAE-based transfer experiments use the exact identifiers
\texttt{PixArt-alpha/\allowbreak PixArt-XL-2-1024-MS}
\citep{chen2024pixartalpha} and
\texttt{diffusers/\allowbreak FLUX.2-dev-bnb-4bit}
\citep{bfl2025flux2}. Both use their checkpoint-provided Diffusers pipeline
and scheduler. All pretrained components remain frozen.
Table~\ref{tab:implementation_settings} collects the resolution, precision,
sampling, and loop settings associated with these checkpoints.

\begin{table*}[t]
\centering
\begingroup
\small
\setlength{\tabcolsep}{3pt}
\begin{tabular}{l|cccccccc}
\toprule
\textbf{Generator} & \textbf{Resolution} & \textbf{Precision} &
\textbf{Steps} & \textbf{Dense range} &
\shortstack{\textbf{Sparse range}\\\textbf{/ ratio}} &
\textbf{$K/\lambda_{\mathrm{loop}}$} &
\textbf{Progress} &
\shortstack{\textbf{$g_{\mathrm{lg}}$}\\\textbf{dense/sparse}} \\
\midrule
Scale-RAE DiT2.4B & $224^2$ & FP32 & 25 & $[12,27]$ &
$[12,19]/70\%$ & $4/1$ &
$[0,0.5]$ & $2.8/\text{--}$ \\
Scale-RAE DiT9.8B & $224^2$ & FP32 & 25 & $[12,27]$ &
$[12,27]/50\%$ & $4/1$ &
$[0,0.5]$ & $6.0/\text{--}$ \\
RAE-SigLIP2-B proxy & $256^2$ & BF16 & 50 & $[15,22]$ &
$[15,22]/50\%$ & $4/1$ &
$[0,0.5]$ & $9.0/6.0$ \\
PixArt-alpha & $1024^2$ & BF16 & 20 & $[14,19]$ & -- & $4/1$ &
$[0,0.5]$ & $3.0/\text{--}$ \\
FLUX.2 & $1024^2$ & BF16/4-bit & 28 & $[32,37]$ & -- & $4/1$ &
$[0,0.5]$ & $3.0/\text{--}$ \\
\bottomrule
\end{tabular}
\endgroup
\caption{Complete loop configurations used for the main and transfer
comparisons. Precision denotes parameter or weight precision. A dash indicates
that Sparse Token Loop was not evaluated for that generator. The reported
Loop Guidance scales are listed as dense/sparse; a dash indicates that the
corresponding guided variant was not reported in the main comparison.
Unguided loop variants are equivalent to $g_{\mathrm{lg}}=1$. The
Scale-RAE DiT9.8B TF32 matmul policy is described in the text, and controlled
Scale-RAE ablations vary only the named factor.}
\label{tab:implementation_settings}
\end{table*}

\subsection{Runtime Configuration and Timing}\leavevmode\par

All reported generation runs use NVIDIA H100 GPUs. Each resident
Scale-RAE Qwen7B/DiT9.8B generation worker is allocated two H100 GPUs and
loads the checkpoint across them with \texttt{device\_map=auto}; this is layer
placement rather than tensor parallelism. All other image generators use one
H100 per generation worker; FLUX.2 additionally enables model CPU offload.
The pinned Scale-RAE, RAE-SigLIP2-B proxy, and PixArt-alpha environment records
Python 3.10, CUDA 12.4, and PyTorch 2.5.1. The separate FLUX.2 overlay records
PyTorch 2.6.0 with CUDA 12.4 and pins its source-installed Diffusers revision;
each official evaluator runs in its own pinned environment.
The generation launcher enables TF32 for CUDA and cuDNN matmuls and sets high
float32 matmul precision. Scale-RAE nevertheless retains FP32 model
parameters and an FP32 diffusion head; the RAE-SigLIP2-B proxy and PixArt
use BF16 autocast, and FLUX.2 uses BF16 compute with 4-bit model weights.
Generation time is measured with a monotonic wall clock around the model
generation call followed by device synchronization. It includes denoiser and
sampler computation but excludes model loading, prompt tokenization, image
decoding, and artifact writing. Timing tables average the per-image records
specified in their captions, and absolute times are compared only within a
matched hardware and software source.

\subsection{Detailed Hyperparameters for Main-Paper Tables and Figures}

The entries below follow their first appearance in the main paper.
Checkpoint-level resolution, precision, sampling, and default loop settings
are collected in Table~\ref{tab:implementation_settings}.

\paragraph{Figure~1: Qualitative examples and
quality--efficiency trade-off.}
The left examples use Scale-RAE DiT9.8B with 25 outer steps, layers
$[12,27]$, $K=4$, $\lambda_{\mathrm{loop}}=1$, and progress $[0,0.5]$.
Dense Token Loop uses $g_{\mathrm{lg}}=1$ and Loop Guidance uses
$g_{\mathrm{lg}}=6.0$, under prompt- and seed-matched generation with seed
42. The right plots use DiT2.4B. Their no-loop points use
$(N,K)\in\{(15,1),(20,1),(25,1),(35,1),(50,1)\}$; Dense Token Loop uses
$(10,8)$, $(15,4)$, and $(25,4)$; and Sparse Token Loop uses $(25,4)$.
All looped points use layers $[12,19]$,
$\lambda_{\mathrm{loop}}=1$, and progress $[0,0.5]$. The sparse point uses
random cached-complement routing with $\rho_{\mathrm{sel}}=0.70$.

\paragraph{Figure~2: Method overview.}
This is an algorithmic schematic rather than the output of an experimental
run, so it has no checkpoint-, seed-, sampler-, or dataset-specific
hyperparameters.

\paragraph{Table~1: Main comparison.}
For Scale-RAE DiT2.4B, Dense Token Loop uses layers $[12,27]$ and Sparse
Token Loop uses layers $[12,19]$ with random cached-complement routing and
$\rho_{\mathrm{sel}}=0.70$. For DiT9.8B, both variants use $[12,27]$, and
the sparse variant uses random cached-complement routing with
$\rho_{\mathrm{sel}}=0.50$. For the RAE-SigLIP2-B proxy, both variants use
$[15,22]$, and the sparse variant uses random cached-complement routing with
$\rho_{\mathrm{sel}}=0.50$. The generator-level outer steps and precision
follow Table~\ref{tab:implementation_settings}. All loop variants use $K=4$,
$\lambda_{\mathrm{loop}}=1$, progress $[0,0.5]$, and the unguided setting
$g_{\mathrm{lg}}=1$. Scale-RAE uses model guidance level 1.0, whereas the
RAE-SigLIP2-B proxy disables classifier-free guidance.

\paragraph{Table~2: Loop Guidance.}
The loop paths retain the preceding dense and sparse configurations.
DiT2.4B Dense Token Loop uses $g_{\mathrm{lg}}=2.8$, DiT9.8B Dense Token
Loop uses $g_{\mathrm{lg}}=6.0$, and the RAE-SigLIP2-B proxy uses
$g_{\mathrm{lg}}=9.0$ for Dense Token Loop and $g_{\mathrm{lg}}=6.0$ for
Sparse Token Loop. Each baseline row is the ordinary denoiser path, denoted
by $g_{\mathrm{lg}}=0$ in the table.

\paragraph{Figure~3: Guidance and loop-count
sweeps.}
Both panels use Scale-RAE DiT2.4B with 25 outer steps, layers $[12,27]$,
$\lambda_{\mathrm{loop}}=1$, and progress $[0,0.5]$. The guidance panel
fixes $K=4$ and varies $g_{\mathrm{lg}}$ over the complete 20-point grid
reported in Table~\ref{tab:loop_guidance}. The loop-count panel fixes
$g_{\mathrm{lg}}=1$ and varies
$K\in\{1,\ldots,8\}$; Table~\ref{tab:k_sweep} reports the complete values.

\paragraph{Figure~4: Layer range and active
progress.}
Both panels use Scale-RAE DiT2.4B with 25 outer steps, $K=4$,
$\lambda_{\mathrm{loop}}=1$, and $g_{\mathrm{lg}}=1$. The layer-range panel
fixes progress $[0,0.5]$ and varies only $[a,b]$; the progress panel fixes
$[a,b]=[12,27]$ and varies only the active interval. The exact candidate
ranges and intervals are listed in Tables~\ref{tab:block_sweep} and
\ref{tab:timestep_sweep}, respectively.

\paragraph{Table~3: Sampling-Progress
Gating.}
This DiT2.4B comparison fixes 25 outer steps, layers $[12,27]$, $K=4$,
$\lambda_{\mathrm{loop}}=1$, and $g_{\mathrm{lg}}=1$, and changes only the
active interval among no loop, $[0,1]$, and $[0,0.5]$.

\noindent\begin{minipage}{\columnwidth}
\paragraph{Table~4: Sparsifiers.}
The Scale-RAE DiT2.4B rows fix 25 outer steps, layers $[12,19]$, $K=4$,
$\lambda_{\mathrm{loop}}=1$, progress $[0,0.5]$, and
$g_{\mathrm{lg}}=1$. They compare random, condition-aware, and
attention-aware routing at $\rho_{\mathrm{sel}}\in\{0.50,0.70\}$ with
cached complement reuse. Random routing samples the selected-token set
uniformly; condition-aware routing selects the strongest
condition-dependent AdaLN responses; and attention-aware routing selects the
strongest gated attention-output responses within each latent-grid row. The
RAE-SigLIP2-B rows fix 50 outer steps, layers $[15,22]$, $K=4$,
$\lambda_{\mathrm{loop}}=1$, progress $[0,0.5]$, and
$g_{\mathrm{lg}}=6.0$. They compare random 50\% routing with an
image/condition split in which image tokens form the selected set and
conditioning tokens form the complement; both use cached complement reuse.
\end{minipage}\par\smallskip

\paragraph{Table~5: VAE transfer.}
PixArt-alpha uses 20 outer steps, layers $[14,19]$, $K=4$,
$\lambda_{\mathrm{loop}}=1$, progress $[0,0.5]$, BF16 autocast, and model
guidance 4.5. FLUX.2 uses 28 outer steps, layers $[32,37]$, the same loop
settings, BF16 compute, 4-bit weights, model CPU offload, and model guidance
4.0. Dense Token Loop is unguided, and the Loop Guidance row uses
$g_{\mathrm{lg}}=3.0$ for both generators.

\FloatBarrier
\noindent\begin{minipage}{\columnwidth}
\section{Complete Scale-RAE Ablations}

All ablations in this subsection use Scale-RAE Qwen1.5B/DiT2.4B.
The main-paper sampling-progress, loop-count, and Loop Guidance figures use
the same pinned GenEval and DPG-Bench evaluators. Their shared no-loop row uses
25 outer sampling steps. The tables below report the complete scores and
per-image generation times.
\paragraph{Outer-sampling-step choice.}
Although the Scale-RAE baselines with 35 and 50 outer sampling steps score
higher, we use 25 outer sampling steps as the experimental default to halve
the sampling budget relative to the native 50-point schedule.
Table~\ref{tab:sampler_steps} reports the official scores and per-image times
for all four sampler settings.

% The native Scale-RAE schedule contains 50 timestep points (49 update
% intervals); we use the conventional 50-step label in the paper.
% Aggregate scores and matched-hardware timings were audited before release.
\begin{center}
\begin{minipage}{\columnwidth}
\centering
\begingroup
\small
\setlength{\tabcolsep}{5pt}
\begin{tabular}{r|ccc}
\toprule
\shortstack{\textbf{Outer sampling}\\\textbf{steps}} &
\textbf{GenEval$\uparrow$} & \textbf{DPG-Bench$\uparrow$} &
\shortstack{\textbf{Time per}\\\textbf{image}\\\textbf{(s)$\downarrow$}} \\
\midrule
50 & 0.4742 & \textbf{0.7738} & 2.4032 \\
35 & \textbf{0.4770} & 0.7673 & 1.7575 \\
\textbf{25} & 0.4471 & 0.7656 & 1.3054 \\
15 & 0.4418 & 0.7645 & 0.8562 \\
\bottomrule
\end{tabular}
\endgroup
\captionof{table}{Twenty-five outer sampling steps are the default efficiency setting.
All rows are unlooped Scale-RAE results. For 15, 35, and 50 outer sampling
steps, times average 1,618 generation records per setting; all rows use the
same pinned evaluators.}
\label{tab:sampler_steps}
\end{minipage}
\end{center}
\end{minipage}

\noindent\begin{minipage}{\columnwidth}
\paragraph{Sampling-progress sweep.}
Table~\ref{tab:timestep_sweep} gives the complete comparison of
loop-active sampling-progress intervals.

\begin{center}
\begin{minipage}{\columnwidth}
\centering
\begingroup
\small
\setlength{\tabcolsep}{2pt}
\begin{tabular}{lr|ccc}
\toprule
\textbf{Progress} & \textbf{Calls} & \textbf{GenEval$\uparrow$} &
\shortstack{\textbf{DPG-}\\\textbf{Bench$\uparrow$}} &
\shortstack{\textbf{Time/image}\\\textbf{(s)$\downarrow$}} \\
\midrule
baseline & 0 & 0.4471 & 0.7656 & 1.3054 \\
\midrule
0--20\% & 5 & 0.5082 & 0.7942 & 1.6416 \\
20--40\% & 5 & 0.5099 & 0.7914 & 1.6408 \\
40--60\% & 5 & 0.4820 & 0.7816 & 1.6392 \\
60--80\% & 5 & 0.4723 & 0.7761 & 1.6389 \\
80--100\% & 5 & 0.4705 & 0.7684 & 1.6385 \\
0--30\% & 8 & 0.5182 & 0.7946 & 1.8407 \\
18--48\% & 7 & 0.5043 & 0.7933 & 1.7731 \\
35--65\% & 7 & 0.4961 & 0.7885 & 1.7727 \\
52--82\% & 7 & 0.4773 & 0.7794 & 1.7728 \\
70--100\% & 8 & 0.4687 & 0.7750 & 1.8401 \\
0--40\% & 10 & 0.5309 & 0.7999 & 1.9771 \\
15--55\% & 10 & 0.5286 & 0.7972 & 1.9758 \\
30--70\% & 9 & 0.4993 & 0.7916 & 1.9080 \\
45--85\% & 10 & 0.4984 & 0.7831 & 1.9759 \\
60--100\% & 10 & 0.4784 & 0.7780 & 1.9767 \\
0--50\% & 13 & 0.5422 & \textbf{0.8007} & 2.1812 \\
25--75\% & 13 & 0.4907 & 0.7956 & 2.1856 \\
50--100\% & 13 & 0.4946 & 0.7831 & 2.1849 \\
0--100\% & 25 & \textbf{0.5597} & 0.7987 & 2.9957 \\
\bottomrule
\end{tabular}
\endgroup
\captionof{table}{Earlier and wider loop-active sampling-progress intervals perform best.
Progress runs from the initial noise state (0\%) to the data endpoint (100\%).
All scores use the same pinned evaluators, and times are matched-H100
generation seconds per image.}
\label{tab:timestep_sweep}
\end{minipage}
\end{center}
\end{minipage}

\noindent\begin{minipage}{\columnwidth}
\paragraph{Loop-count sweep.}
Table~\ref{tab:k_sweep} reports the complete range of loop counts from
$K=1$ through $K=8$.

\begin{center}
\begin{minipage}{\columnwidth}
\centering
\begingroup
\small
\setlength{\tabcolsep}{4pt}
\begin{tabular}{r|ccc}
\toprule
\textbf{Loop Count} & \textbf{GenEval$\uparrow$} &
\textbf{DPG-Bench$\uparrow$} &
\shortstack{\textbf{Time per}\\\textbf{image}\\\textbf{(s)$\downarrow$}} \\
\midrule
1 & 0.4471 & 0.7656 & 1.3054 \\
\midrule
2 & 0.5180 & 0.7977 & 1.5949 \\
3 & 0.5278 & 0.7962 & 1.8881 \\
4 & \textbf{0.5422} & 0.8007 & 2.1812 \\
5 & 0.5322 & 0.7999 & 2.4744 \\
6 & 0.5308 & 0.7996 & 2.7676 \\
7 & 0.5268 & 0.7990 & 3.0607 \\
8 & 0.5376 & \textbf{0.8037} & 3.3537 \\
\bottomrule
\end{tabular}
\endgroup
\captionof{table}{Moderate loop counts give the best quality--time balance.
GenEval peaks at $K=4$, DPG-Bench peaks at $K=8$, and
measured time per image increases nearly linearly.}
\label{tab:k_sweep}
\end{minipage}
\end{center}
\end{minipage}

\noindent\begin{minipage}{\columnwidth}
\paragraph{Layer-range sweep.}
Table~\ref{tab:block_sweep} reports the complete layer-range ablation.

\begin{center}
\begin{minipage}{\columnwidth}
\centering
\begingroup
\small
\setlength{\tabcolsep}{3pt}
\begin{tabular}{ll|cc}
\toprule
\textbf{\#Layers} & \textbf{Layer range} & \textbf{GenEval$\uparrow$} &
\textbf{DPG-Bench$\uparrow$} \\
\midrule
0 & baseline & 0.4471 & 0.7656 \\
\midrule
4 & 28--31 & 0.5090 & 0.7807 \\
4 & 18--21 & 0.4990 & 0.7889 \\
4 & 10--13 & 0.4617 & 0.7726 \\
4 & 0--3 & 0.4605 & 0.7650 \\
8 & 16--23 & 0.5282 & 0.8009 \\
8 & 24--31 & 0.5421 & 0.7850 \\
8 & 12--19 & 0.5113 & 0.7942 \\
8 & 8--15 & 0.4611 & 0.7767 \\
8 & 0--7 & 0.4592 & 0.7695 \\
16 & 12--27 & \textbf{0.5422} & 0.8007 \\
16 & 16--31 & 0.5407 & 0.8049 \\
16 & 4--19 & 0.5249 & 0.7978 \\
16 & 0--15 & 0.4735 & 0.7773 \\
32 & 0--31 & 0.5395 & \textbf{0.8087} \\
\bottomrule
\end{tabular}
\endgroup
\captionof{table}{Mid-to-late layer ranges work best. Both metrics are
computed with the same pinned evaluators.}
\label{tab:block_sweep}
\end{minipage}
\end{center}
\end{minipage}

\begin{figure*}[t]
\centering
\includegraphics[width=0.94\textwidth]{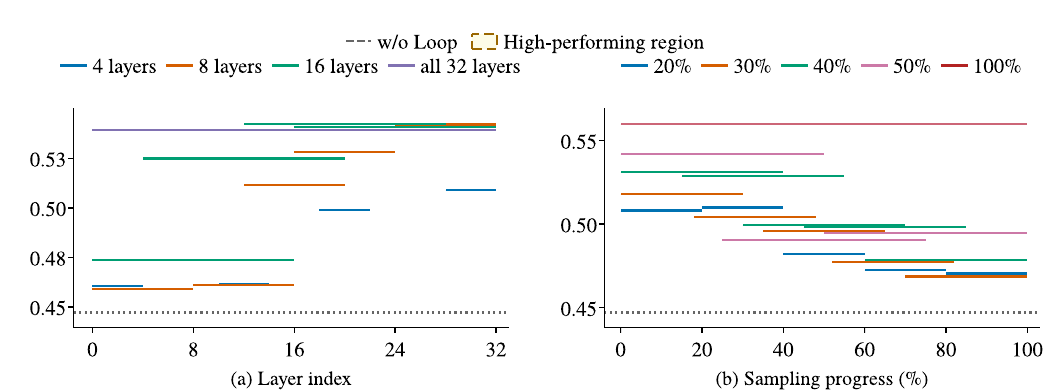}
\caption{GenEval counterparts for the loop layer range and loop-active
sampling-progress interval ablations on Scale-RAE (DiT2.4B). Mid-to-late layer
ranges and earlier sampling-progress intervals give the strongest gains,
consistent with the DPG-Bench results in the main paper.}
\label{fig:window_ablation_geneval_supp}
\end{figure*}

\paragraph{GenEval layer-range and sampling-progress panels.}
Figure~\ref{fig:window_ablation_geneval_supp} provides the GenEval counterparts
to the main-paper DPG-Bench ablations.

\noindent\begin{minipage}{\columnwidth}
\paragraph{Loop Guidance sweep.}
Table~\ref{tab:loop_guidance} reports every guidance scale covered by the
complete evaluation.

\begin{center}
\begin{minipage}{\columnwidth}
\centering
\begingroup
\small
\setlength{\tabcolsep}{6pt}
\begin{tabular}{r|cc}
\toprule
\textbf{$g_{\mathrm{lg}}$} & \textbf{GenEval$\uparrow$} &
\textbf{DPG-Bench$\uparrow$} \\
\midrule
0.0 & 0.4471 & 0.7656 \\
\midrule
0.5 & 0.5131 & 0.7895 \\
1.0 & 0.5405 & 0.8011 \\
1.5 & 0.5530 & 0.8065 \\
2.0 & 0.5582 & \textbf{0.8083} \\
2.4 & 0.5577 & 0.8019 \\
2.8 & \textbf{0.5691} & 0.8053 \\
3.2 & 0.5534 & 0.8035 \\
3.6 & 0.5520 & 0.8028 \\
4.0 & 0.5523 & 0.8037 \\
4.5 & 0.5425 & 0.8025 \\
5.0 & 0.5456 & 0.8030 \\
5.5 & 0.5477 & 0.8035 \\
6.0 & 0.5399 & 0.8011 \\
7.0 & 0.5277 & 0.7977 \\
8.0 & 0.5220 & 0.7960 \\
9.0 & 0.5176 & 0.7825 \\
10.0 & 0.5236 & 0.7801 \\
11.0 & 0.5169 & 0.7759 \\
12.0 & 0.5093 & 0.7706 \\
\bottomrule
\end{tabular}
\endgroup
\captionof{table}{Moderate Loop Guidance gives the strongest gains.
GenEval peaks at $g_{\mathrm{lg}}=2.8$, while DPG-Bench peaks at
$g_{\mathrm{lg}}=2.0$; all entries use the same pinned evaluators.}
\label{tab:loop_guidance}
\end{minipage}
\end{center}
\end{minipage}

\FloatBarrier
\section{Other Experimental Results}

\paragraph{RAE-SigLIP2-B layer-range sweep.}
Table~\ref{tab:raev2_transfer_full} reports the complete layer-range sweep
underlying the direct-loop result in the main paper, and
Table~\ref{tab:raev2_lg_full} reports the corresponding Loop Guidance sweep.
We use the DiffusionBench RAE-SigLIP2-B/DDT proxy identified under
Models and Checkpoints.
All variants use 50 outer sampling steps with Euler updates, disabled CFG,
the loop-active sampling-progress interval $s_i\in[0,0.5]$, and dense token
updates.

% Aggregate scores and evaluator bindings were audited before release.
\begin{center}
\begin{minipage}{\columnwidth}
\centering
\begingroup
\small
\setlength{\tabcolsep}{3pt}
\begin{tabular}{lr|cc}
\toprule
\textbf{Layer range} & \textbf{Loop Count} &
\textbf{GenEval$\uparrow$} & \textbf{DPG-Bench$\uparrow$} \\
\midrule
No loop & 1 & 0.3829 & 0.7131 \\
\midrule
0--3 & 4 & 0.0000 & 0.2382 \\
6--9 & 4 & 0.3789 & 0.7135 \\
12--15 & 4 & 0.3764 & 0.7174 \\
18--21 & 4 & \textbf{0.3873} & 0.7172 \\
24--27 & 4 & 0.0644 & 0.4093 \\
\midrule
0--7 & 4 & 0.0021 & 0.2379 \\
5--12 & 4 & 0.3678 & 0.7195 \\
10--17 & 4 & 0.3806 & 0.7170 \\
15--22 & 4 & 0.3795 & \textbf{0.7226} \\
20--27 & 4 & 0.0685 & 0.4048 \\
\bottomrule
\end{tabular}
\endgroup
\captionof{table}{One middle RAE layer range improves both primary benchmarks.
Several other middle ranges raise DPG-Bench while slightly lowering GenEval,
and extreme early and late ranges collapse; all entries are absolute standard
scores from the pinned GenEval and DPG-Bench evaluator protocols.}
\label{tab:raev2_transfer_full}
\end{minipage}
\end{center}

\paragraph{RAE-SigLIP2-B Loop Guidance.}
The sweep fixes loop layer range $[15,22]$ with a loop count of $K=4$.
Increasing
$g_{\mathrm{lg}}$ generally strengthens both metrics: $g_{\mathrm{lg}}=9$
maximizes GenEval, while $g_{\mathrm{lg}}=7.5$ gives the highest DPG-Bench
score.

% Aggregate scores and evaluator bindings were audited before release.
\begin{center}
\begin{minipage}{\columnwidth}
\centering
\begingroup
\small
\setlength{\tabcolsep}{5pt}
\begin{tabular}{r|cc}
\toprule
\textbf{$g_{\mathrm{lg}}$} & \textbf{GenEval$\uparrow$} &
\textbf{DPG-Bench$\uparrow$} \\
\midrule
0.00 & 0.3829 & 0.7131 \\
\midrule
1.50 & 0.3787 & 0.7271 \\
2.25 & 0.3962 & 0.7286 \\
3.00 & 0.3892 & 0.7350 \\
3.75 & 0.3983 & 0.7397 \\
4.50 & 0.4120 & 0.7393 \\
5.25 & 0.4094 & 0.7495 \\
6.00 & 0.4217 & 0.7468 \\
6.75 & 0.4220 & 0.7483 \\
7.50 & 0.4274 & \textbf{0.7526} \\
8.25 & 0.4308 & 0.7487 \\
9.00 & \textbf{0.4532} & 0.7521 \\
\bottomrule
\end{tabular}
\endgroup
\captionof{table}{Loop Guidance substantially improves both DiffusionBench
RAE-SigLIP2-B scores. The $g_{\mathrm{lg}}=0$ row is the no-loop prediction;
all entries use the same pinned evaluators.}
\label{tab:raev2_lg_full}
\end{minipage}
\end{center}

\FloatBarrier
\subsection{Additional Qualitative Results}
\label{sec:loop_correction_supp}

\begin{figure*}[t]
\centering
\begingroup
\small
\setlength{\tabcolsep}{0pt}
\begin{tabular}{@{}p{0.118\textwidth}*{3}{p{0.1255\textwidth}}p{0.127\textwidth}*{3}{p{0.1255\textwidth}}@{}}
& \centering\textbf{w/o Loop}\par
& \centering\textbf{Dense Token Loop}\par
& \centering\textbf{Loop Guidance}\par
&
& \centering\textbf{w/o Loop}\par
& \centering\textbf{Dense Token Loop}\par
& \centering\textbf{Loop Guidance}\par
\end{tabular}\par
\vspace{0.1em}
\includegraphics[width=\textwidth]{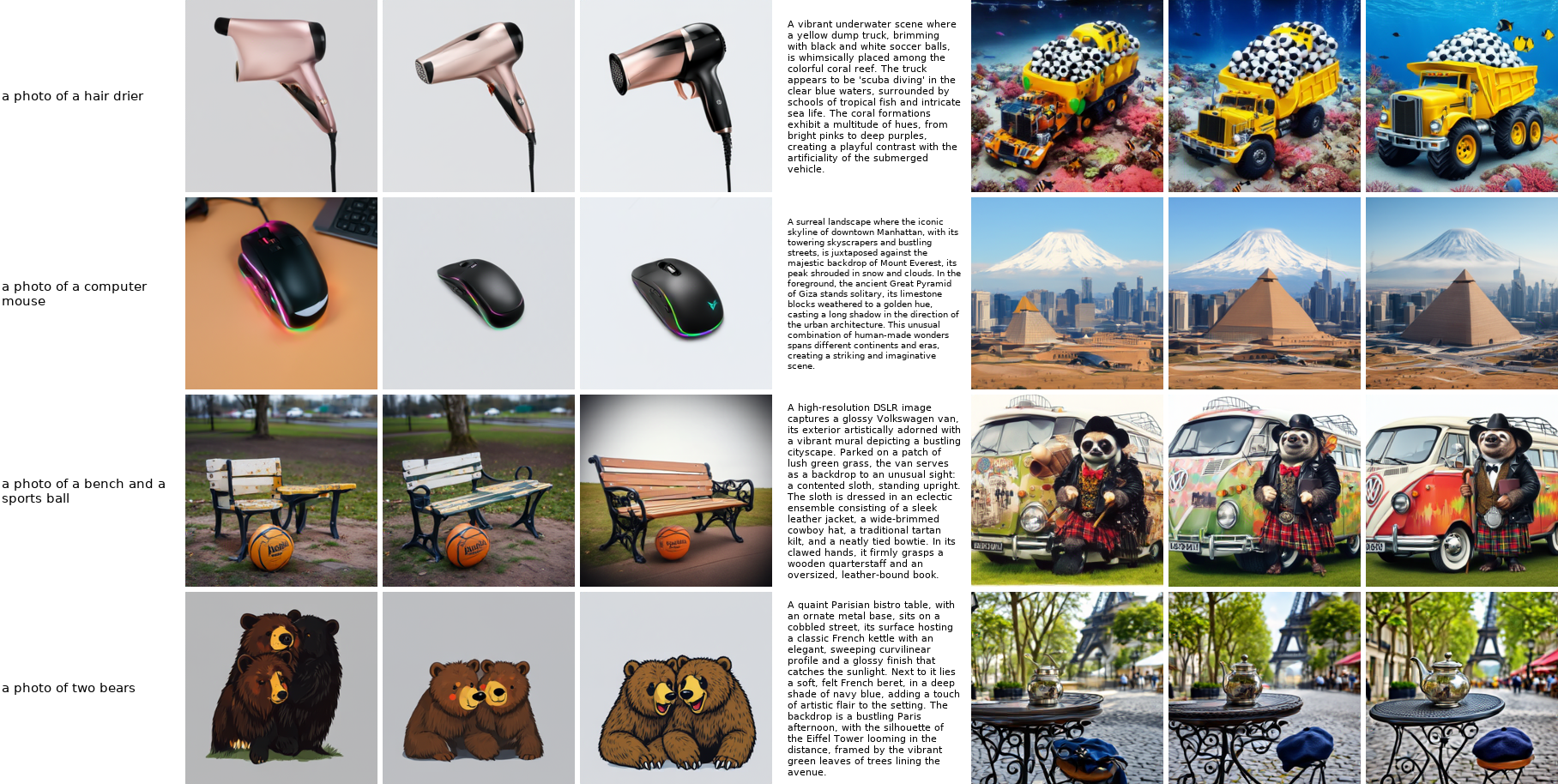}
\vspace{0.1em}
\begin{tabular}{@{}p{0.5\textwidth}p{0.5\textwidth}@{}}
\centering\textit{GenEval}\par & \centering\textit{DPG-Bench}\par
\end{tabular}
\endgroup
\caption{Additional prompt-matched qualitative comparisons on Scale-RAE
(DiT9.8B). Dense Token Loop uses loop layer range $[12,27]$, a loop count of
$K=4$, and loop-active sampling-progress interval $s_i\in[0,0.5]$; Loop
Guidance uses $g_{\mathrm{lg}}=6.0$.}
\label{fig:qualitative_correction_supp}
\end{figure*}

\paragraph{Subject-level corrections.}
Manual review of prompt-matched pairs provides a preliminary characterization
of the correction effect. On GenEval and DPG-Bench, whose prompts are dominated
by a single subject or a pair of subjects, looping most consistently corrects
subject-level appearance: malformed or incomplete shapes become well formed,
spurious parts and duplicated features are removed, and attributes such as
color, material, and style are more cleanly bound to their intended entity
instead of bleeding into the background or a neighboring object. The sample as
a whole therefore looks closer to the prompt intent, which is consistent with
the GenEval and DPG-Bench score gains. Figure~\ref{fig:qualitative_correction_supp}
provides additional prompt-matched examples.

\FloatBarrier

\paragraph{GenEval2 findings.}
On GenEval2, whose prompts stress object counts, positional relations, and
deliberately unconventional phrasing, Table~\ref{tab:geneval2_supp} reports
official soft-TIFA-style geometric-mean scores over 800 prompts
\citep{hu2023tifa}. Dense Token Loop changes the score from $0.1801$ to
$0.1736$, while Loop Guidance reaches $0.1441$. Thus, neither variant improves
the aggregate GenEval2 score. Nevertheless, manual review of matched triplets
(w/o Loop, Dense Token Loop, and Loop Guidance) reveals a structured pattern,
summarized in Figure~\ref{fig:geneval2_correction_supp}. On prompts with one or
two subjects, looping gives the same correction benefit as on GenEval and
DPG-Bench: objects are better formed and attributes are bound more cleanly. On
counting prompts, the requested object count is usually inherited from the
no-loop sample rather than corrected. On crowded multi-object prompts,
individual objects can remain better formed even when their global arrangement
does not improve. Looped samples also sometimes weaken or remove the requested
background while purifying the foreground subject.

% The GenEval2 aggregate was audited before release.
\begin{center}
\begin{minipage}{\columnwidth}
\centering
\begingroup
\small
\setlength{\tabcolsep}{5pt}
\begin{tabular}{l|c}
\toprule
\textbf{Method} & \textbf{GenEval2$\uparrow$} \\
\midrule
w/o Loop & \textbf{0.1801} \\
\midrule
Dense Token Loop & 0.1736 \\
Loop Guidance & 0.1441 \\
\bottomrule
\end{tabular}
\endgroup
\captionof{table}{GenEval2 results on Scale-RAE (DiT9.8B) over 800 prompts.
Dense Token Loop uses loop layer range $[12,27]$, a loop count of $K=4$, and
loop-active sampling-progress interval $s_i\in[0,0.5]$; Loop Guidance
additionally uses $g_{\mathrm{lg}}=6.0$.}
\label{tab:geneval2_supp}
\end{minipage}
\end{center}

\begin{figure*}[t]
\centering
\begingroup
\setlength{\tabcolsep}{3pt}
\begin{tabular}{cc}
\includegraphics[width=0.48\textwidth]{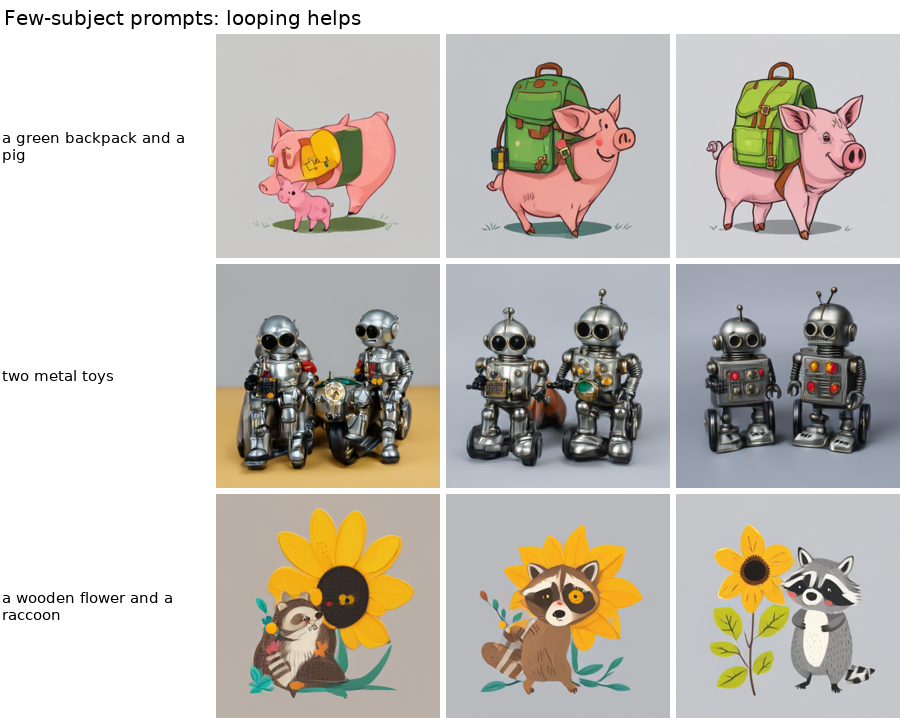} &
\includegraphics[width=0.48\textwidth]{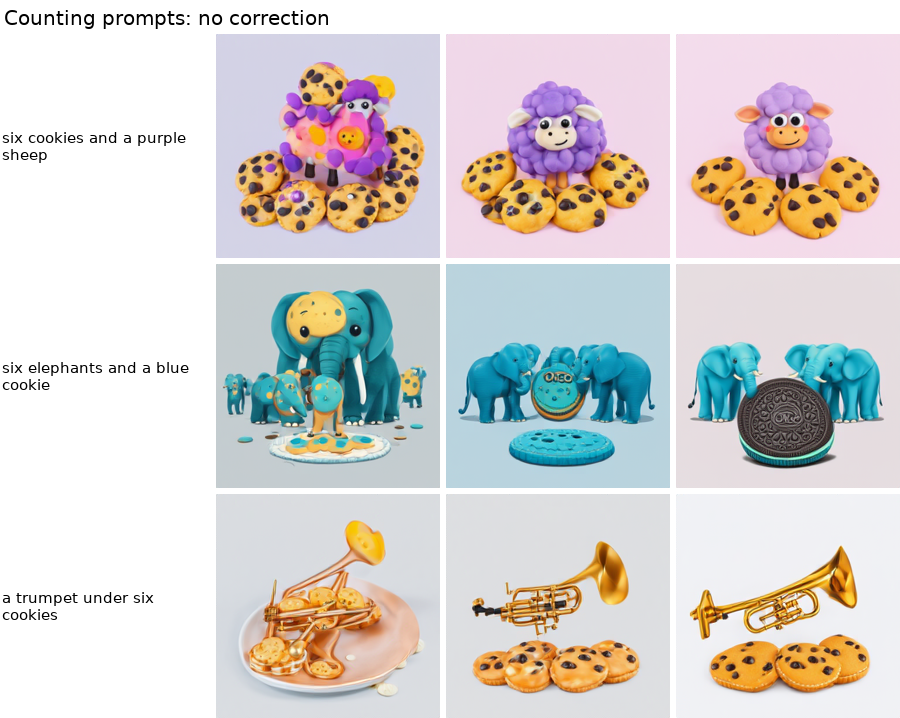} \\
\includegraphics[width=0.48\textwidth]{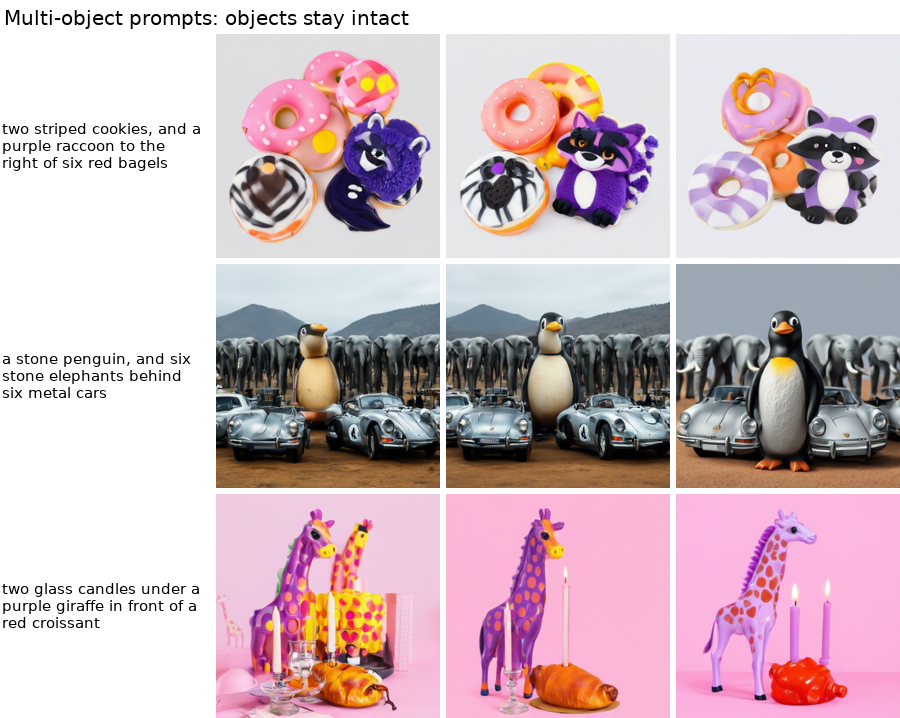} &
\includegraphics[width=0.48\textwidth]{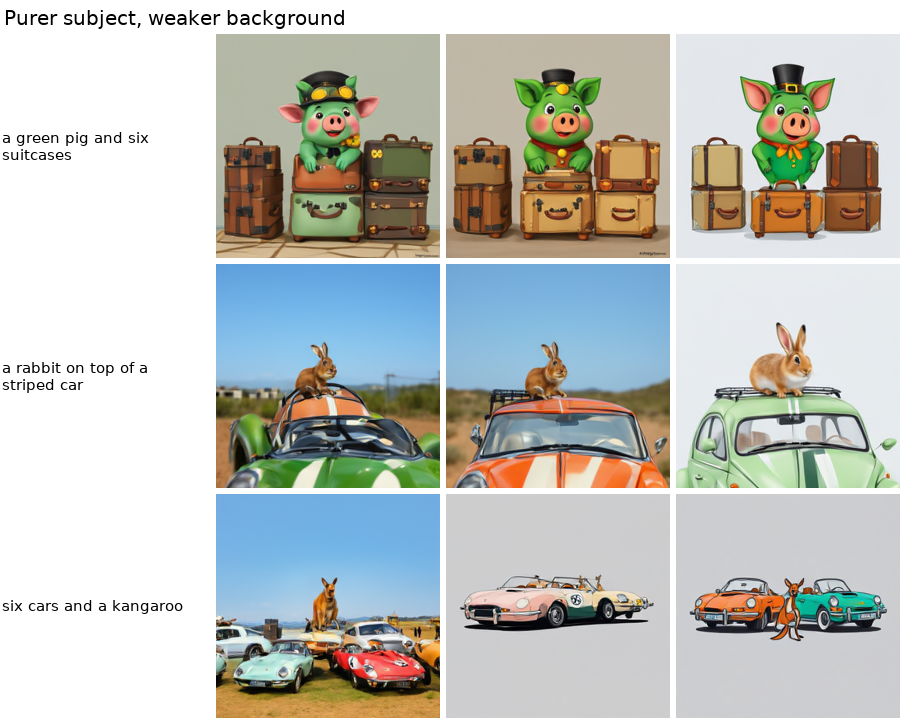}
\end{tabular}
\endgroup
\caption{GenEval2 qualitative correction study on Scale-RAE
(DiT9.8B). Each row inside a panel shows the prompt followed by w/o Loop, Dense
Token Loop (loop layer range $[12,27]$, a loop count of $K=4$, and sampling
progress $s_i\in[0,0.5]$), and Loop Guidance
($g_{\mathrm{lg}}=6.0$). Top left: few-subject prompts improve. Top right:
counting prompts are not corrected. Bottom left: objects in crowded
multi-object prompts stay intact. Bottom right: subjects become purer with
weaker backgrounds.}
\label{fig:geneval2_correction_supp}
\end{figure*}
\FloatBarrier

\section{Reproducibility}

\paragraph{Code and Data Supplement.}
The anonymous Code and Data Supplement contains the backend-neutral loop
runtime, adapters for Scale-RAE, the public RAE-SigLIP2-B/DDT proxy,
PixArt-alpha, and FLUX.2, exact JSON configurations, complete benchmark prompt
snapshots, benchmark generation/export/aggregation utilities, CPU-only tests,
and the aggregate values used by the paper tables. The archive excludes
pretrained weights, generated answers and image grids, per-prompt evaluator
outputs, credentials, caches, logs, and machine-specific paths. After
installing the environment described in \texttt{ENVIRONMENT.md}, the commands
\texttt{python scripts/validate\_release.py} and
\texttt{python -m unittest discover tests} validate the archive without
loading a generator. \texttt{REPRODUCE.md} then gives the backend integration
and benchmark procedure. Model checkpoints and official benchmark inputs are
obtained separately using the public identifiers in \texttt{configs/}.

\paragraph{Composable method and hyperparameter interface.}
The released resolver treats the token-loop base and Loop Guidance as
orthogonal choices. \texttt{-{}-token-loop dense|sparse} selects Dense Token
Loop or Sparse Token Loop, while \texttt{-{}-loop-guidance} may be composed
with either base and never changes its token-loop type.
Tables~\ref{tab:code_hyperparameters} and
\ref{tab:sparse_hyperparameters} list the complete shared interface for
adjustable paper controls. Scale-RAE and RAEv2 expose both token-loop bases
with or without Loop Guidance. The PixArt-$\alpha$ and FLUX.2 paper presets
expose Dense Token Loop with or without Loop Guidance because Sparse Token
Loop was not evaluated for those backends; unsupported combinations are
rejected instead of being silently relabeled.

\begin{table*}[!t]
\centering
\begingroup
\small
\setlength{\tabcolsep}{2.0pt}
\begin{tabular}{p{0.16\textwidth}|p{0.22\textwidth}|p{0.22\textwidth}|p{0.34\textwidth}}
\toprule
\textbf{Paper control} & \textbf{JSON field} &
\textbf{Shared CLI override} & \textbf{Semantics and accepted values} \\
\midrule
Outer sampler steps $N$ &
\shortstack[l]{\texttt{generation.num\_}\\\texttt{inference\_steps}} &
\texttt{-{}-outer-steps N} &
Positive integer; number of outer denoising steps. \\
Loop count $K$ &
\texttt{num\_loops} &
\texttt{-{}-loop-count K} &
Positive integer; inner updates at each active outer step. \\
Total loop strength $\lambda_{\mathrm{loop}}$ &
\texttt{lambda\_value} &
\texttt{-{}-lambda-loop VALUE} &
Positive scalar; the runtime applies $\lambda_{\mathrm{loop}}/K$ per inner round. \\
Loop-active sampling interval &
\texttt{start\_frac}, \texttt{end\_frac} &
\texttt{-{}-loop-active START END} &
$0\leq\texttt{START}\leq\texttt{END}\leq1$; both endpoints are active. \\
Loop layer range $[a,b]$ &
\texttt{block\_indices} &
\texttt{-{}-loop-layers FIRST LAST} &
Inclusive, contiguous, ascending, nonnegative layer indices. \\
Token-loop base &
\texttt{token\_operator} &
\texttt{-{}-token-loop none|dense|sparse} &
Ordinary path, Dense Token Loop, or Sparse Token Loop. \\
Loop Guidance switch &
\texttt{loopguidance\_enabled} &
\texttt{-{}-loop-guidance} &
Boolean modifier; valid with either Dense or Sparse Token Loop. \\
Loop Guidance scale $g_{\mathrm{lg}}$ &
\texttt{loopguidance\_weight} &
\shortstack[l]{\texttt{-{}-loop-guidance-}\\\texttt{weight VALUE}} &
$0\leq g_{\mathrm{lg}}\leq18$; requires \texttt{-{}-loop-guidance}. \\
\bottomrule
\end{tabular}
\endgroup
\caption{Released code-facing core hyperparameter interface. The JSON files in
\texttt{configs/} are the exact paper presets. A CLI override is applied to a
copy and sets \texttt{paper\_preset=false} in the resolved record, preventing a
custom run from being mistaken for a reported configuration.}
\label{tab:code_hyperparameters}
\end{table*}

\begin{table*}[!t]
\centering
\begingroup
\small
\setlength{\tabcolsep}{2.0pt}
\begin{tabular}{p{0.16\textwidth}|p{0.22\textwidth}|p{0.22\textwidth}|p{0.34\textwidth}}
\toprule
\textbf{Sparse control} & \textbf{JSON field} &
\textbf{Shared CLI override} & \textbf{Semantics and accepted values} \\
\midrule
Selector &
\texttt{selector} &
\texttt{-{}-selector NAME} &
\texttt{all}, \texttt{random}, \texttt{condition\_aware},
\texttt{attention\_aware}, or \texttt{image\_condition\_split}. \\
Selected-token fraction $\rho_{\mathrm{sel}}$ &
\texttt{selection\_ratio} &
\shortstack[l]{\texttt{-{}-selection-ratio}\\\texttt{RATIO}} &
$0<\rho_{\mathrm{sel}}\leq1$ when required by the sparse selector. \\
Sparse-routing seed &
\texttt{selector\_seed} &
\texttt{-{}-selector-seed SEED} &
Integer seed for reproducible token routing. \\
Eligible token domain &
\texttt{token\_domain} &
\shortstack[l]{\texttt{-{}-token-domain}\\\texttt{DOMAIN}} &
\texttt{backend\_default}, \texttt{all\_tokens},
\texttt{image\_prefix\_only}, or \texttt{image\_vs\_condition}. \\
\bottomrule
\end{tabular}
\endgroup
\caption{Sparse Token Loop routing controls. Unsupported backend/selector
combinations fail validation before model loading.}
\label{tab:sparse_hyperparameters}
\end{table*}
\FloatBarrier

\paragraph{Evaluator and artifact integrity.}
Every reported Scale-RAE ablation row is rescored from its bound image grid with
the same pinned evaluators. GenEval uses the pristine evaluator without
pre-upscaling, while DPG-Bench fixes the CSV parser, one image per prompt,
native evaluation resolution, and the local mPLUG checkpoint. Incomplete,
superseded, or evaluator-incompatible artifacts are excluded from the tables
so evaluator drift is not interpreted as a method improvement.

\end{document}